\documentclass{article}
\usepackage{ijcai24}

\usepackage{times}
\usepackage{soul}
\usepackage{url}
\usepackage[hidelinks]{hyperref}
\usepackage[utf8]{inputenc}
\usepackage[small]{caption}
\usepackage{graphicx}
\usepackage{amsmath}
\usepackage{amsthm}
\usepackage{booktabs}
\usepackage[noend]{algpseudocode}
\usepackage{algorithm}
\usepackage{algorithmicx}
\algnewcommand{\LeftComment}[1]{\Statex \(\triangleright\) #1}
\usepackage[switch]{lineno}

\usepackage{amssymb,amsfonts}
\usepackage[hang,flushmargin]{footmisc}
\usepackage{multirow}
\usepackage{tabularx}
\newcolumntype{R}{>{\raggedleft\arraybackslash}X}
\newcolumntype{C}{>{\centering\arraybackslash}X}
\newcolumntype{L}{>{\hsize=4\hsize}X}
\newcolumntype{Z}{>{\hsize=1\hsize}C}
\newcolumntype{J}{>{\hsize=1\hsize}C}
\newcolumntype{B}{>{\hsize=0.889\hsize}R}
\newcolumntype{Y}{>{\hsize=1\hsize}R}
\usepackage{bm}
\usepackage{textcomp}
\usepackage[table]{xcolor}
\usepackage{siunitx}
\usepackage{tikz}
\usetikzlibrary{shapes,backgrounds,shapes.geometric,arrows.meta}
\usepackage{pgfplots}
\usepackage{standalone}

\usepackage{mathrsfs}

\usepackage{jabbrv}

\DeclareMathOperator*{\argmin}{arg\,min}
\definecolor{ROME_color}{RGB}{25,125,50}

\def\BibTeX{{\rm B\kern-.05em{\sc i\kern-.025em b}\kern-.08em
    T\kern-.1667em\lower.7ex\hbox{E}\kern-.125emX}}

\newcommand{\splitatcommas}[1]{%
  \begingroup
  \begingroup\lccode`~=`, \lowercase{\endgroup
    \edef~{\mathchar\the\mathcode`, \penalty0 \noexpand\hspace{0pt plus 1em}}%
  }\mathcode`,="8000 #1%
  \endgroup
}

\title{ROME: Robust Multi-Modal Density Estimator}

\author{
Anna M\'esz\'aros
\and
Julian F. Schumann\and
Javier Alonso-Mora\and \\
Arkady Zgonnikov\And
Jens Kober
\affiliations
Cognitive Robotics, TU Delft, Netherlands 
\emails
\{A.Meszaros, J.F.Schumann, J.AlonsoMora, A.Zgonnikov, J.Kober\}@tudelft.nl
}

\begin{document}
\maketitle

\begin{abstract}
The estimation of probability density functions is a fundamental problem in science and engineering. 
However, common methods such as kernel density estimation (KDE) have been demonstrated to lack robustness, while more complex methods have not been evaluated in multi-modal estimation problems. 
In this paper, we present ROME (RObust Multi-modal Estimator), a non-parametric approach for density estimation which addresses the challenge of estimating multi-modal, non-normal, and highly correlated distributions. ROME utilizes clustering to segment a multi-modal set of samples into multiple uni-modal ones and then combines simple KDE estimates obtained for individual clusters in a single multi-modal estimate. We compared our approach to state-of-the-art methods for density estimation as well as ablations of ROME, showing that it not only outperforms established methods but is also more robust to a variety of distributions. Our results demonstrate that ROME can overcome the issues of over-fitting and over-smoothing exhibited by other estimators.
\end{abstract}

\section{Introduction}
Numerous processes are non-deterministic by nature, from geological and meteorological occurrences, biological activities, as well as the behavior of living beings. Estimating the underlying probability density functions (PDFs) of such processes enables a better understanding of them and opens possibilities for probabilistic inference regarding future developments. Density estimation is instrumental in many applications including classification, anomaly detection, and evaluating probabilistic AI models, such as generative adversarial networks~\cite{goodfellow2020generative}, variational autoencoders~\cite{wei2020variations}, normalizing flows~\cite{kobyzev2020normalizing}, and their numerous variations.

\begin{figure}[t]
    \centering
    \includegraphics[width=\linewidth]{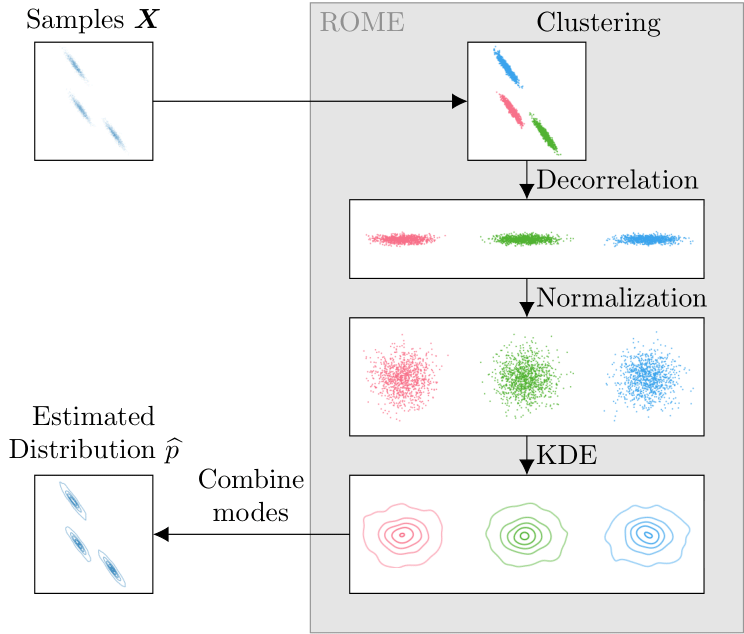}
    \caption{ROME takes samples from unknown distributions and estimates their densities to enable further downstream applications.}
    \label{fig:pipeline}
\end{figure}

When probabilistic models are trained on multi-modal data, they are often evaluated using simplistic metrics (e.g., mean squared error (MSE) between the predicted and ground truth samples).
However, such simplistic metrics are unsuited for determining how well a predicted distribution corresponds to the underlying distribution, as they do not capture the fit to the whole distribution. For example, the lowest MSE value between true and predicted samples could be achieved by accurate predictions of the mean of the true underlying distribution whereas potential differences in variance or shape of the distribution would not be penalized.
This necessitates more advanced metrics that evaluate the match between the model and the (potentially multi-modal) data. For instance, negative log-likelihood (NLL), Jensen-Shannon divergence (JSD), and expected calibration error (ECE) can be used to evaluate how well the \textit{full distribution} of the data is captured by the learned models~\cite{xu2019survey,mozaffari2020deep,rasouli2020deep}. However, most data-driven models represent the learned distribution implicitly, only providing individual samples and not the full distribution as an output. This complicates the comparison of the model output to the ground-truth data distributions since the above metrics require distributions, not samples, as an input. Practically, this can be addressed by estimating the predicted probability density based on samples generated by the model.

Simple methods like Gaussian mixture models (GMM), kernel density estimation (KDE)~\cite{deisenroth2020mathematics}, and k-nearest neighbors (kNN)~\cite{loftsgaarden1965nonparametric} are commonly used for estimating probability density functions. These estimators however rely on strong assumptions about the underlying distribution, and can thereby introduce biases or inefficiencies in the estimation process. For example, problems can arise when encountering multi-modal, non-normal, and highly correlated distributions (see Section~\ref{sec:related_work}).
While more advanced methods such as manifold Parzen windows (MPW)~\cite{vincent2002manifold} and vine copulas (VC)~\cite{nagler2016evading} exist, they have not been thoroughly tested on such problems, which raises questions about their performance.

To overcome these limitations, we propose a novel density estimation approach: RObust Multi-modal Estimator (ROME). ROME employs a non-parametric clustering approach to segment potentially multi-modal distributions into separate uni-modal ones (Figure~\ref{fig:pipeline}). These uni-modal sub-distributions are then estimated using a downstream probability density estimator (such as KDE). We test our proposed approach against a number of existing density estimators in three simple two-dimensional benchmarks designed to evaluate a model's ability to successfully reproduce multi-modal, highly-correlated, and non-normal distributions. Finally, we test our approach in a real-world setting using a distribution over future trajectories of human pedestrians created based on the Forking Paths dataset~\cite{liang2020garden}. 

\section{Related Work}
\label{sec:related_work}
The most common class of density estimators are so-called Parzen windows~\cite{parzen1962estimation}, which estimate the density through an aggregation of parametric probability density functions. A number of common methods use this approach, with KDE being a common non-parametric method~\cite{silverman2018density}. Provided a type of kernel -- which is oftentimes a Gaussian but can be any other type of distribution -- KDE places the kernels around each sample of a data distribution and then sums over these kernels to get the final density estimation over the data. This method is often chosen as it does not assume any underlying distribution type~\cite{silverman2018density}. However, if the underlying distribution is highly correlated, then the common use of a single isotropic kernel function can lead to over-smoothing~\cite{vincent2002manifold,wang2009fast}. 
Among multiple approaches for overcoming this issue~\cite{wang2009fast,gao2022adaptive}, especially noteworthy is the MPW approach~\cite{vincent2002manifold}. 
It uses a unique anisotropic kernel for every datapoint, estimated based on the correlation of the k-nearest neighbors of each sample. However, it has not been previously tested in high-dimensional benchmarks, which is especially problematic as the required memory scales quadratically with the dimensionality of the problem.

Another common subtype of Parzen windows are GMMs~\cite{deisenroth2020mathematics}, which assume that the data distribution can be captured through a weighted sum of Gaussian distributions. The parameters of the Gaussian distributions -- also referred to as components -- are estimated through expected likelihood maximization. Nonetheless, especially for non-normal distributions, one needs to have prior knowledge of the expected number of components to achieve a good fit without over-fitting to the training data or over-smoothing~\cite{mclachlan2014number}.

Besides different types of Parzen windows, a prominent alternative is kNN~\cite{loftsgaarden1965nonparametric} which uses the local density of the $k$ nearest neighbors of every data point to estimate the overall density. 
While this method is non-parametric, it cannot be guaranteed that the resulting distribution will be normalized~\cite{silverman2018density}. 
This could be rectified by using Monte Carlo sampling to obtain a good coverage of the function’s domain and obtain an accurate estimate of the normalization factor, which, however, becomes computationally intractable for high-dimensional distributions.

When it comes to estimating densities characterized by correlations between dimensions, copula-based methods are an often favored approach. 
Copula-based methods decompose a distribution into its marginals and an additional function, called a copula, which describes the dependence between these marginals over a marginally uniformly distributed space. 
The downside of most copula-based approaches is that they rely on choosing an adequate copula function (e.g., Gaussian, Student, or Gumbel) and estimating their respective parameters \cite{joe2014dependence}. 
One non-parametric copula-based density estimator~\cite{bakam2023nonparametric} aims to address this limitation by estimating copulas with the help of superimposed Legendre polynomials. 
While this can achieve good results in estimating the density function, it may become computationally intractable as the distribution's dimensionality increases. 
Another approach involves the use of VC~\cite{nagler2016evading}, which assume that the whole distribution can be described as a product of bivariate Gaussian distributions, thus alleviating the curse of dimensionality. 
Its convergence, however, can only be guaranteed for normal distributions.
Elsewhere~\cite{otneim2017locally}, a similar approach was pursued, with changes such as using logarithmic splines instead of univariate KDEs for estimating the marginals. 
However, both of these approaches are not designed for multi-modal distributions and have not been thoroughly tested on such problems.

\section{RObust Multi-modal Estimator (ROME)}
\begin{algorithm}[t!]
\begin{algorithmic}
\Function{TrainROME}{$\bm{X}$}
\State{\(\triangleright\) \textit{Clustering (OPTICS)}}
\State $\bm{X}_{I,N}, \bm{R}_N \leftarrow$ \Call{ReachabiltyAnalysis}{$\bm{X}$}
\State $\bm{C}, S \leftarrow \{\{1, \hdots, N\}\}, -1.1$
\ForAll{$\epsilon \in \bm{\varepsilon}$}
    \State $\bm{C}_{\epsilon} \leftarrow \text{\textbf{DBSCAN}}(\bm{R}_{I,N}, \epsilon)$
    \State $S_{\epsilon} \leftarrow \text{\textbf{SIL}}(\bm{C}_{\epsilon}, \bm{X}_{I,N})$ 
    \If{$S_{\epsilon} > S$}
        \State $\bm{C}, S \leftarrow \bm{C}_{\epsilon}, S_{\epsilon} $
    \EndIf
\EndFor
\ForAll{$\xi \in \bm{\xi}$}
    \State $\bm{C}_{\xi} \leftarrow \text{$\bm{\xi}$\textbf{-clustering}}(\bm{R}_{I,N}, \xi)$
    \State $S_{\xi} \leftarrow \text{\textbf{SIL}}(\bm{C}_{\xi}, \bm{X}_{I,N})$ 
    \If{$S_{\xi} > S$}
        \State $\bm{C}, S \leftarrow \bm{C}_{\xi}, S_{\xi} $
    \EndIf
\EndFor
\ForAll{$C \in \bm{C}$}
    \State{\(\triangleright\) \textit{Decorrelation}}
    \State $\overline{\bm{x}}_{C} \leftarrow \text{MEAN}(\bm{X}_{C}) $
    \State $\overline{\bm{X}}_{C} \leftarrow \bm{X}_{C} - \overline{\bm{x}}_{C} $
    \State $\bm{R}_{C} \leftarrow PCA(\overline{\bm{X}}_{C})$
    \State{\(\triangleright\) \textit{Normalization}}
    \State $\widetilde{\bm{\Sigma}}_{C} \leftarrow \text{STD}(\overline{\bm{X}}_{C} \bm{R}_{C}^T )$
    \State $\bm{T}_{C} \leftarrow  \bm{R}_{C}^T  \widetilde{\bm{\Sigma}}_{C}^{-1}$
    \State{\(\triangleright\) \textit{PDF Estimation}}
    \State $\widehat{p}_{C} \leftarrow f_{\text{KDE}}\left(\overline{\bm{X}}_{C} \bm{T}_{C}\right)$
\EndFor

\Return $\bm{C}$, $\left\{\widehat{p}_{C}, \overline{\bm{x}}_{C}, \bm{T}_C \,\vert\, C \in \bm{C} \right\}$

\EndFunction \\
\Function{QueryROME}{$\bm{x}$, $\bm{X}$}
\State 
$\bm{C}$, $\left\{\widehat{p}_{C}, \overline{\bm{x}}_{C}, \bm{T}_C \,\vert\, C \in \bm{C} \right\} \leftarrow$ \Call{TrainROME}{$\bm{X}$}
\State $l = \bm{0}$
\ForAll{$C \in \bm{C}$}
    \State $\widehat{\bm{x}} \leftarrow \left(\bm{x} - \overline{\bm{x}}_{C}\right) \bm{T}_{C} $
    \State $l \leftarrow l + \ln(\widehat{p}_C(\widehat{\bm{x}}))+ \ln(\vert C \vert) - \ln{N} + \ln(\vert \det (\bm{T}_C)\vert) $

\EndFor

\Return $\exp (l)$

\EndFunction
\caption{ROME}
\label{algo::ROME}
\end{algorithmic}
\end{algorithm}

The problem of density estimation can be formalized as finding a queryable $\widehat{p} \in \mathcal{P}$, where
\begin{equation*}
\mathcal{P} = \left\{g: \mathbb{R}^{M} \rightarrow \mathbb{R}^+  \, \vert \, \textstyle{\int} \, g(\bm{x}) \, \text{d} \bm{x} = 1\right\},
\end{equation*} 
such that $\widehat{p}$ is close to the non-queryable PDF $p$ underlying the $N$ available $M$-dimensional samples $\bm{X} \in \mathbb{R}^{N \times M}$:  $\bm{X} \sim p$.

A solution to the above problem would be an estimator $f: \mathbb{R}^{N \times M} \rightarrow \mathcal{P}$, resulting in $\widehat{p} = f(\bm{X})$. 
Our proposed estimator $f_{\text{ROME}}$\footnote{Source code: \url{https://github.com/anna-meszaros/ROME}} (Algorithm~\ref{algo::ROME}) is built on top of non-parametric cluster extraction. 
Namely, by separating groups of samples surrounded by areas of low density -- expressing the mode of the underlying distribution -- we reduce the multi-modal density estimation problem to multiple uni-modal density estimation problems for each cluster. 
The distributions within each cluster then become less varied in density or correlation than the full distribution.
Combining this with decorrelation and normalization, the use of established methods such as KDE to estimate probability densities for those uni-modal distributions is now more promising, as problems with multi-modality and correlated modes (see Section~\ref{sec:related_work}) are accounted for.
The multi-modal distribution is then obtained as a weighted average of the estimated uni-modal distributions. 

\subsection{Extracting Clusters}
\label{sec:clustering}
To cluster samples $\bm{X}$, ROME employs the OPTICS algorithm~\cite{ankerst1999optics} that can detect clusters of any shape with varying density using a combination of reachability analysis -- which orders the data in accordance to reachability distances -- followed by clustering the ordered data based on these reachability distances. 

In the first part of the algorithm, the reachability analysis is used to sequentially transfer samples from a set of unincluded samples $\bm{X}_{U,i}$ to the set of included and ordered samples $\bm{X}_{I,i}$, starting with a random sample $\bm{x}_1$ ($\bm{X}_{I,1} = \{\bm{x}_1\}$ and $\bm{X}_{U,1} = \bm{X} \backslash \{\bm{x}_1\}$). 
The samples $\bm{x}_{i+1}$ are then selected at iteration $i$ based on the reachability distance $r$:
\begin{equation}
    \bm{x}_{i+1} = \underset{\bm{x} \in \bm{X}_{U,i}}{\argmin} \, r(\bm{x}, \bm{X}_{I,i}) = \underset{\bm{x} \in \bm{X}_{U,i}}{\argmin} \;\, \underset{\bm{\widetilde{x}} \in \bm{X}_{I,i}}{\min\vphantom{g}} \,  d_r\left(\bm{x}, \bm{\widetilde{x}}\right) \, . \label{eq:reachability}
\end{equation}
This sample is then transferred between sets, with
$\bm{X}_{I,i+1} = \bm{X}_{I,i} \cup \{\bm{x}_{i+1}\}$ and $\bm{X}_{U,{i+1}} = \bm{X}_{U,i} \backslash \{\bm{x}_{i+1}\}$, while expanding the reachability set $\bm{R}_{i+1} = \bm{R}_{i} \cup \left\{r(\bm{x}_{i+1}, \bm{X}_{I,i}) \right\}$ (with $\bm{R}_1 = \left\{\infty \right\}$).
The reachability distance $d_r$ in Equation \eqref{eq:reachability} is defined as
\begin{equation*}
    d_r\left(\bm{x}, \bm{\widetilde{x}}\right) = \max\left\{\left\Vert \bm{x} - \bm{\widetilde{x}} \right\Vert,  \underset{\bm{\widehat{x}} \in \bm{X}\backslash \{\bm{x}\}}{{\min}_{k_{\bm{c}}}}  \left\Vert \bm{x} - \bm{\widehat{x}} \right\Vert  \right\} \, ,
\end{equation*}
where ${\min}_{k_{\bm{c}}}$ is the $k_{\bm{c}}$-smallest value of all the available $\bm{\widehat{x}}$, used to smooth out random local density fluctuations. 
We use 
\begin{equation}
    k_{\bm{c}} = \min\left\{k_{\max}, \max\left\{ k_{\min}, \frac{N M}{\alpha_k}\right\} \right\} \,,
    \label{eq:kC}
\end{equation}
where $k_{\min}$ and $k_{\max}$ ensure there are sufficient but not too many points for this smoothing, while the term $NM / \alpha_k$ adjusts $k_{\bm{c}}$ to the number of samples and dimensions. 

The second part of the OPTICS algorithm -- after obtaining the reachability distances $\bm{R}_N$ and the ordered set $\bm{X}_{I,N}$ -- is the extraction of a set of clusters $\bm{C}$, with clusters $C = \{c_{\min}, \hdots, c_{\max}\} \in \bm{C}$ (with $\bm{X}_{C} = \{\bm{x}_{I,N,j} \, \vert \, j \in C\} \in \mathbb{R}^{\vert C\vert \times M}$).
As the computational cost of creating such a cluster set is negligible compared to the reachability analysis, we can test multiple clusterings generated using two different approaches (with $r_{\text{bound}} = \min \{ r_{N,c_{\min}}, r_{N,c_{\max} + 1}\}$, see Appendix A for further discussion):
\begin{itemize}
    \item First, we use \textbf{DBSCAN}~\cite{ester1996density} for generating the clustering $\bm{C}_\epsilon$ based on an absolute limit $\epsilon$, where a cluster must fulfill the condition:
    \begin{equation}
            r_{N,c} < \epsilon \leq r_{\text{bound}} \; \forall \, c \in C \backslash \{c_{\min} \} \, . \label{eq:DBSCAN}
    \end{equation}
    \item Second, we use $\bm{\xi}$\textbf{-clustering}~\cite{ankerst1999optics} to generate the clustering $\bm{C}_\xi$ based on the proportional limit $\xi$, where a cluster fulfills:
    \begin{equation}
        \xi \leq  1 - \frac{r_{N,c}}{r_{\text{bound}}} \; \forall \, c \in C \backslash \{c_{\min} \} \, . \label{eq:xi}
    \end{equation}
\end{itemize}  
In both cases, each prospective cluster $C$ also has to fulfill the condition $\vert C \vert \geq 2$. However, it is possible that not every sample can be fit into a cluster fulfilling the conditions above. These samples are then kept in a separate noise cluster $C_{\text{noise}}$ that does not have to fulfill those conditions ($C_{\text{noise}} \in \bm{C}_\epsilon$ or $C_{\text{noise}} \in \bm{C}_\xi$ respectively). 
Upon generating multiple sets of clusters $\bm{C}_\epsilon$ ($\epsilon \in \bm{\varepsilon}$) and $\bm{C}_\xi$ ($\xi \in \bm{\xi}$), we select the final set of clusters $\bm{C}$ that achieves the highest silhouette score\footnote{The silhouette score measures the similarity of each object to its own cluster's objects compared to the other clusters' objects.} $S = \text{\textbf{SIL}}(\bm{C}, \bm{X}_{I,N}) \in [-1,1]$~\cite{rousseeuw1987silhouettes}.
The clustering then allows us to use PDF estimation methods on uni-modal distributions.

\subsection{Feature Decorrelation}
\label{sec:decorrelation}
In much of real-life data, such as the distributions of a person's movement trajectories, certain dimensions of the data are likely to be highly correlated.
Therefore, the features in each cluster $C \in \bm{C}$ should be decorrelated using a rotation matrix $\bm{R}_{C} \in \mathbb{R}^{M\times M}$. In ROME, $\bm{R}_{C}$ is found using principal component analysis (PCA)~\cite{wold1987principal} on the cluster's centered samples $\overline{\bm{X}}_{C} = \bm{X}_{C} - \overline{\bm{x}}_{C}$ ($\overline{\bm{x}}_{C}$ is the cluster's mean value). An exception are the noise samples in $C_{\text{noise}}$, which are not decorrelated (i.e., $\bm{R}_{C_\text{noise}} =~\bm{I}$). One can then get the decorrelated samples $\bm{X}_{\text{PCA}, C}$:
\begin{equation*}
    \bm{X}_{\text{PCA}, C}^T=\bm{R}_{C} \overline{\bm{X}}_{C}^T \, .
\end{equation*}

\subsection{Normalization}
\label{sec:normalization}
After decorrelation, we use the matrix $\widetilde{\bm{\Sigma}}_{C} \in \mathbb{R}^{M\times M}$ to normalize $\bm{X}_{\text{PCA}, C}$: 
\begin{equation*}
    \widehat{\bm{X}}_{C} = \bm{X}_{\text{PCA}, C}  \left(\widetilde{\bm{\Sigma}}_{C}\right)^{-1} = \overline{\bm{X}}_{C}  \bm{R}_C^T \left(\widetilde{\bm{\Sigma}}_{C}\right)^{-1} = \overline{\bm{X}}_{C}  \bm{T}_{C} \,.
\end{equation*}
Here, $\widetilde{\bm{\Sigma}}_{C}$ is a diagonal matrix with the entries $\widetilde{\sigma}_{C}^{(m)}$. To avoid over-fitting to highly correlated distributions, we introduce a regularization with a value $\sigma_{\min}$ (similar to~\cite{vincent2002manifold}) that is applied to the empirical standard deviation $\sigma_{\text{PCA}, C}^{(m)} = \sqrt{\mathbb{V}_m\left( \bm{X}_{\text{PCA}, C}\right)}$ ($\mathbb{V}_m$ is the variance along feature $m$) of each rotated feature $m$ (with $\bm{C}' = \bm{C}\backslash \{C_{\text{noise}}\}$):
\begin{equation*}
    \widetilde{\sigma}_{C}^{(m)} = \begin{cases}
        \max\left\{  \sum\limits_{\mathscr{C} \in \bm{C}'}
        \frac{\sqrt{\mathbb{V}_m\left(\bm{X}_{\mathscr{C}}\right) }}{\vert \bm{C}\vert - 1}  , \sigma_{\min} \right\} & C = C_{\text{noise}} \\
        \left( 1 - \frac{\sigma_{\min}}{\underset{m}{\max} \; \sigma_{\text{PCA}, C}^{(m)}}\right) \sigma_{\text{PCA}, C}^{(m)}+ \sigma_{\min} & \text{otherwise}
    \end{cases}.
\end{equation*}

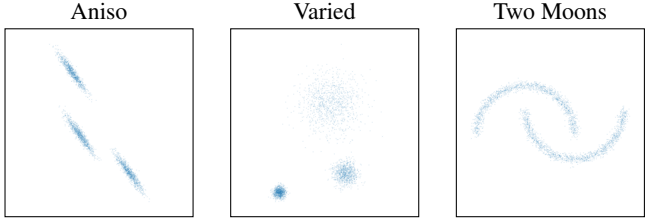
\begin{figure}[!t]
    \centering
    \begin{tikzpicture}
        \begin{scope}[xshift = 0cm]
            \node[above right, inner sep = 0] at (0,0) {\includegraphics[width = 2.5cm]{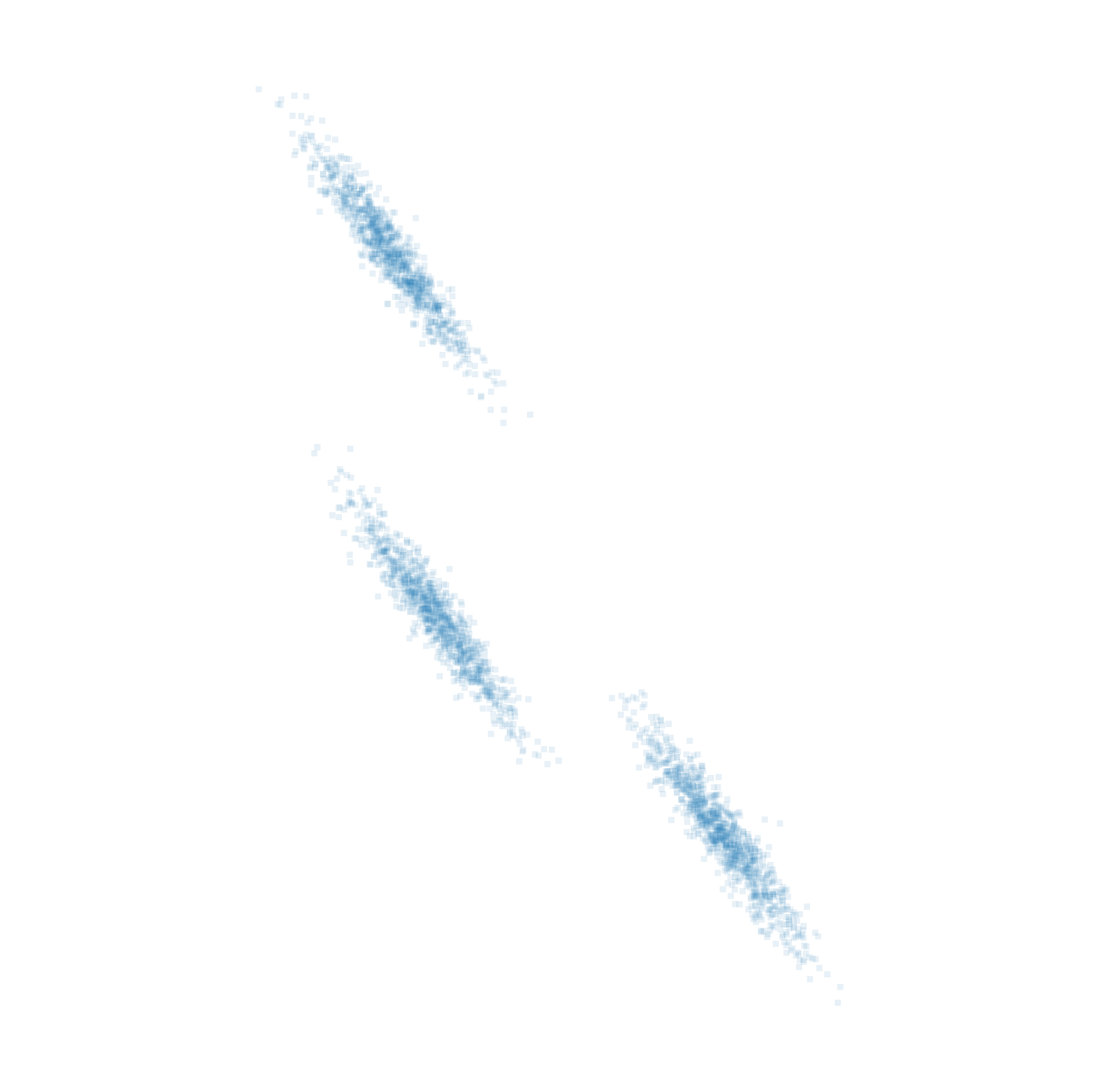}};
            \draw[black] (0,0) rectangle (2.5,2.5);
            \node[above, black, font = \small] at (1.25,2.5) {Aniso};
        \end{scope}
        \begin{scope}[xshift = 3cm]
            \node[above right, inner sep = 0] at (0,0) {\includegraphics[width = 2.5cm]{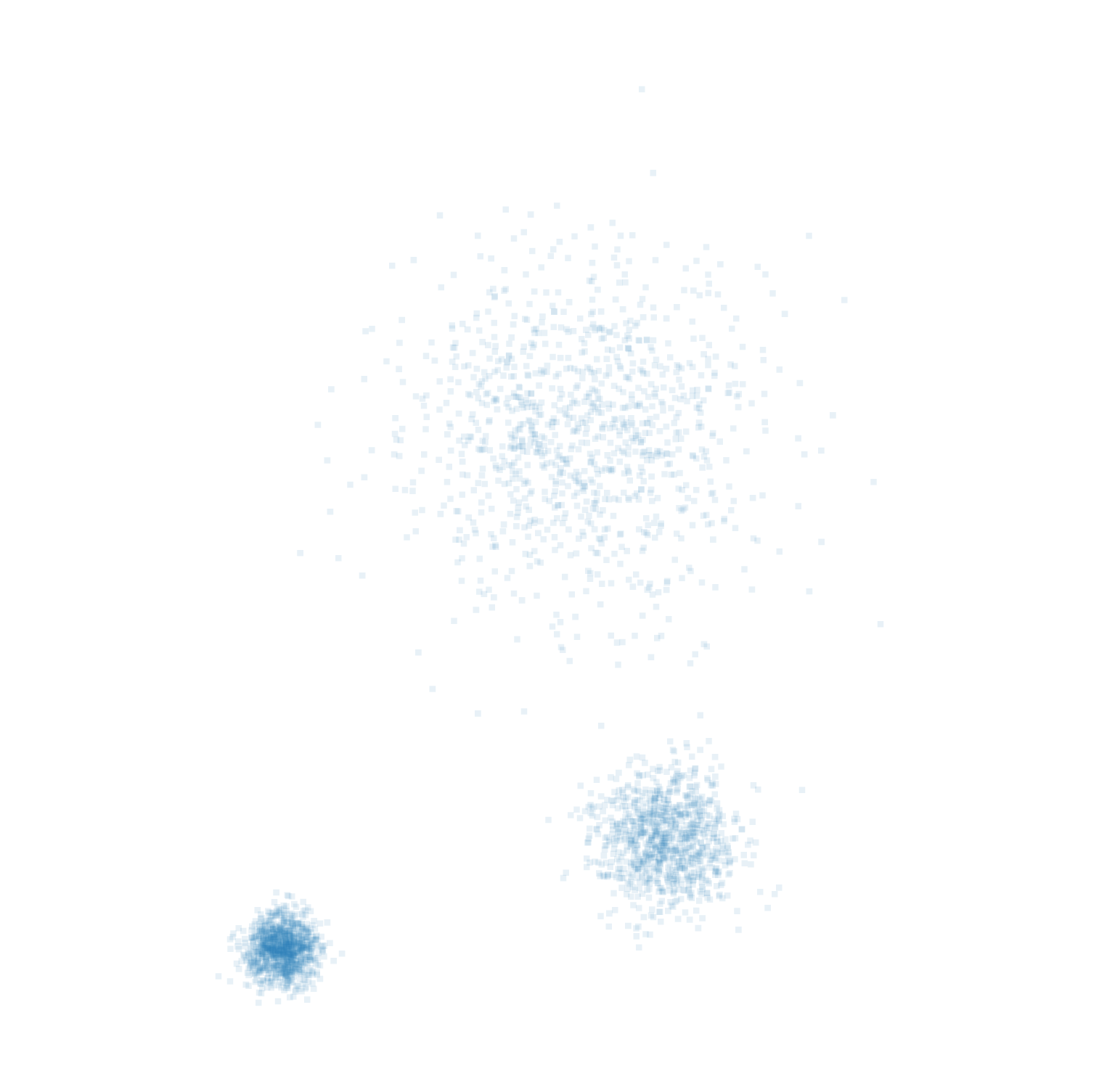}};
            \draw[black] (0,0) rectangle (2.5,2.5);
            \node[above, black, font = \small] at (1.25,2.5) {Varied};
        \end{scope}
        \begin{scope}[xshift = 6cm]
            \node[above right, inner sep = 0] at (0,0) {\includegraphics[width = 2.5cm]{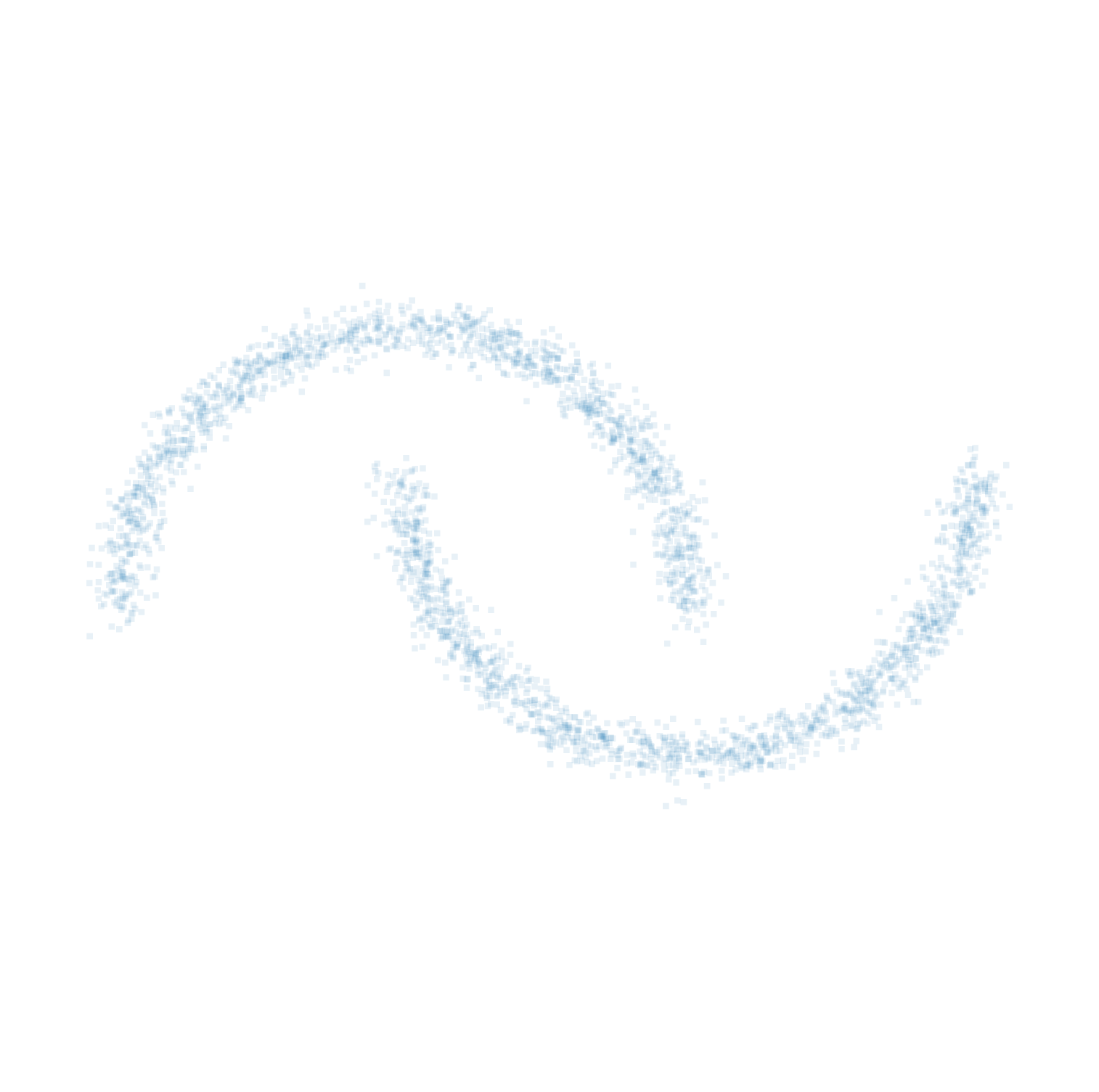}};
            \draw[black] (0,0) rectangle (2.5,2.5);
            \node[above, black, font = \small] at (1.25,2.5) {Two Moons};
        \end{scope}
    \end{tikzpicture}
    \vspace{-4mm}
    \caption{Samples from the two-dimensional synthetic distributions used for evaluating different PDF estimators.}
    \vspace{-2mm}
    \label{fig:2Ddistributions}
\end{figure}

\subsection{Estimating the Probability Density Function}
\label{sec:pdf}
Taking the transformed data (decorrelated and normalized), ROME fits the Gaussian KDE $f_{\text{KDE}}$ on each separate cluster $C$ as well as the noise samples $C_{\text{noise}}$. For a given bandwidth $b_C$ for data in cluster $C$, this results in a partial PDF $\widehat{p}_{C}$.
\begin{equation*}
    \widehat{p}_{C}\left(\widehat{\bm{x}}\right) = f_{\text{KDE}}\left(\widehat{\bm{X}}_{C}\right)\left(\widehat{\bm{x}}\right) = \frac{1}{\vert C \vert} \sum\limits_{\bm{\chi} \in \widehat{\bm{X}}_{C}} \mathcal{N}\left(\widehat{\bm{x}} \vert \bm{\chi} , b_C \bm{I}\right) \,.
\end{equation*}
The bandwidth $b_{C}$ is set using Silverman's rule~\cite{silverman2018density}:
\begin{equation*}
    b_{C} = \left(\frac{M + 2}{4} n_C\right)^{- \frac{1}{M + 4}} \,, \, n_C = \begin{cases} 1 & C = C_{\text{noise}} \\ \vert C \vert & \text{else} \end{cases} \,.
\end{equation*}

To evaluate the density function $\widehat{p} = f_{\text{ROME}}(\bm{X})$ for a given sample $\bm{x}$, we take the weighted averages of each cluster's $\widehat{p}_{C}$:
\begin{equation*}\begin{aligned}
    \widehat{p}\left(\bm{x}\right) = \sum\limits_{C\in\bm{C}} \frac{\vert C\vert }{N}\,
    \widehat{p}_{C}\left(\left(\bm{x} - \overline{\bm{x}}_{C}\right) \bm{T}_{C}\right)  \vert \det\left(\bm{T}_{C}\right)\vert \, .
\end{aligned}\end{equation*}
Here, the term $\vert C \vert / N$ is used to weigh the different distributions of each cluster with the size of each cluster, so that each sample is represented equally. As the different KDEs $\widehat{p}_{C}$ are fitted to the transformed samples, we apply them not to the original sample $\bm{x}$, but instead apply the identical transformation used to generate those transformed samples, using $\widehat{p}_{C}\left(\left(\bm{x} - \overline{\bm{x}}_{C}\right)\right)$. To account for the change in density within a cluster $C$ introduced by the transformation $\bm{T}_{C}$, we use the factor $\vert \det\left(\bm{T}_{C}\right)\vert $.
\begin{figure}[!t]
    \centering
    \begin{tikzpicture}
        \node[above right, inner sep = 0] at (0,0) {\includegraphics[width = 5cm]{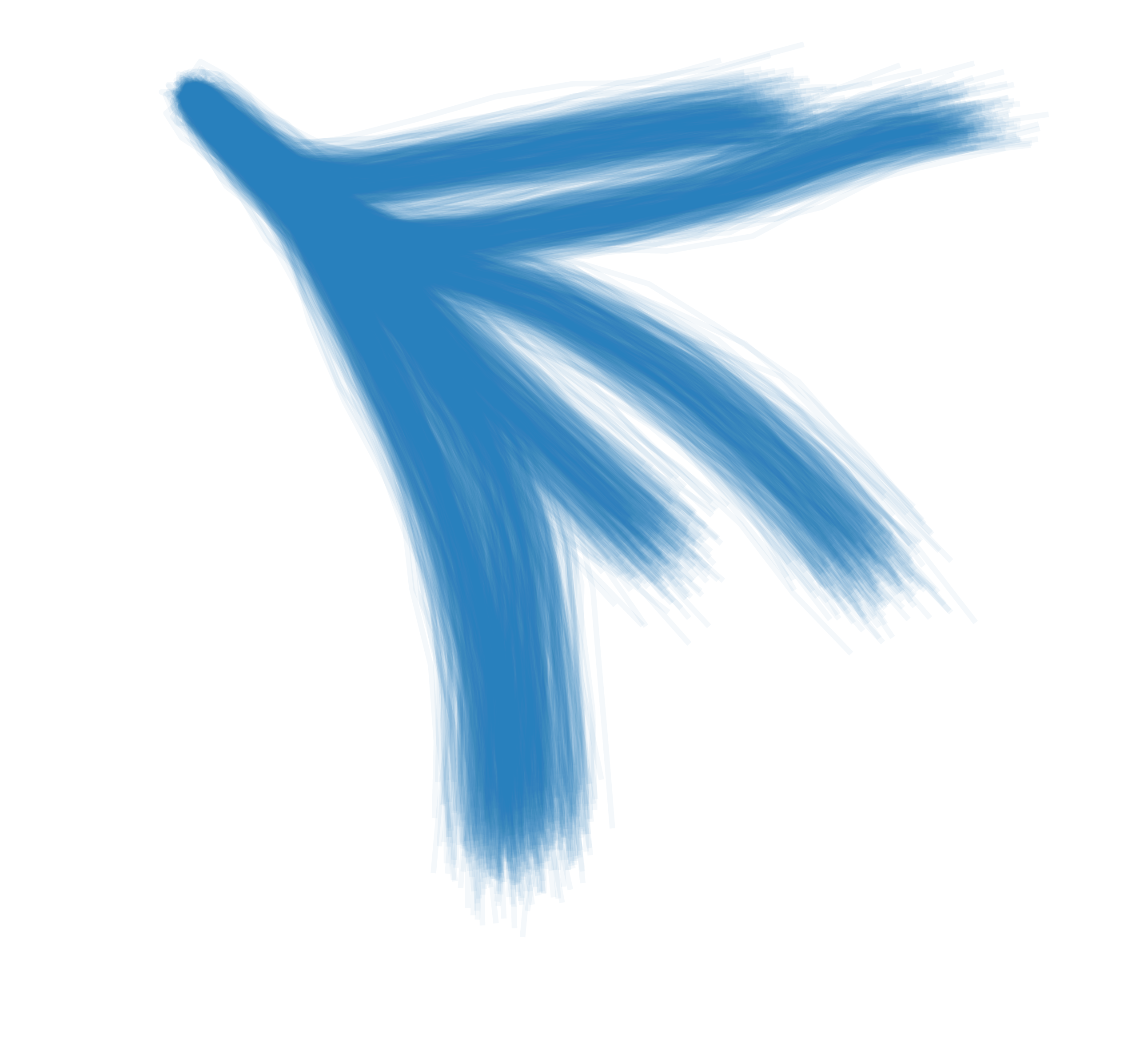}};
        \begin{axis}[ 
            at = {(0cm, 0cm)}, 
            width = 5cm, 
            height = 4.500cm, 
            scale only axis = true, 
            axis lines = left, 
            xmin = 3, 
            xmax = 8, 
            xtick = {3, 4, 5, 6, 7}, 
            xticklabels = {3, 4, 5, 6, 7}, 
            x tick label style = {font=\scriptsize}, 
            xmajorgrids = false, 
            ymin = 4, 
            ymax = 8.5, 
            ytick = {4, 5, 6, 7, 8}, 
            yticklabels = {4, 5, 6, 7, 8}, 
            y tick label style = {font=\scriptsize, rotate = 90}, 
            ymajorgrids = false, 
        ] 
        \end{axis}
        \node[below, black, font = \small] at (2.5,-0.4) {$x$ [$m$]};
        \node[above, black, font = \small, rotate = 90] at (-0.4, 2.25) {$y$ [$m$]};
    \end{tikzpicture}
    \vspace{-2mm}
    \caption{Samples from the multi-modal pedestrian trajectory distribution~\protect\cite{liang2020garden} used for evaluating different PDF estimators. Trajectories span $12$ timesteps recorded at \SI{2.5}{\hertz}.}
    \vspace{-2mm}
    \label{fig:bimodal}
\end{figure}
\begin{table*}[t!]
\centering
\begin{tabular}{l | r@{\hskip 1pt}l @{\hskip 2mm}  r@{\hskip 1pt}l @{\hskip 2mm} r@{\hskip 1pt}l | r@{\hskip 1pt}l @{\hskip 2mm} r@{\hskip 1pt}l @{\hskip 2mm} r@{\hskip 1pt}l | r@{\hskip 1pt}l @{\hskip 2mm} r@{\hskip 1pt}l @{\hskip 2mm} r@{\hskip 1pt}l }
\toprule[1pt] 
 & \multicolumn{6}{c|}{$D_\text{JS} \downarrow_{0}^{1}$} & \multicolumn{6}{c|}{$\widehat{W} \rightarrow 0$} & \multicolumn{6}{c}{$\widehat{L} \uparrow$}  \\
Distrib. & \multicolumn{2}{c}{ROME} & \multicolumn{2}{c}{MPW} & \multicolumn{2}{c|}{VC} & \multicolumn{2}{c}{ROME} & \multicolumn{2}{c}{MPW} & \multicolumn{2}{c|}{VC} & \multicolumn{2}{c}{ROME} & \multicolumn{2}{c}{MPW} & \multicolumn{2}{c}{VC}  \\ \midrule[1pt]
\small Aniso & \scriptsize $0.010$ & $\scriptscriptstyle{\pm 0.001}$ & \scriptsize $0.026$ & $\scriptscriptstyle{\pm 0.002}$ & \scriptsize $\underline{0.005}$ & $\scriptscriptstyle{\pm 0.001}$ & \scriptsize $\underline{-0.13}$ & $\scriptscriptstyle{\pm 0.31}$ & \scriptsize $-0.60$ & $\scriptscriptstyle{\pm 0.13}$ & \scriptsize $1.91$ & $\scriptscriptstyle{\pm 0.91}$ & \scriptsize $\underline{-2.53}$ & $\scriptscriptstyle{\pm 0.02}$ & \scriptsize $-2.57$ & $\scriptscriptstyle{\pm 0.02}$ & \scriptsize $-3.19$ & $\scriptscriptstyle{\pm 0.02}$ \\ \midrule 
\small Varied & \scriptsize $0.011$ & $\scriptscriptstyle{\pm 0.001}$ & \scriptsize $0.025$ & $\scriptscriptstyle{\pm 0.002}$ & \scriptsize $\underline{0.008}$ & $\scriptscriptstyle{\pm 0.001}$ & \scriptsize $\underline{-0.13}$ & $\scriptscriptstyle{\pm 0.20}$ & \scriptsize $-0.49$ & $\scriptscriptstyle{\pm 0.11}$ & \scriptsize $1.27$ & $\scriptscriptstyle{\pm 0.53}$ & \scriptsize $\underline{-4.10}$ & $\scriptscriptstyle{\pm 0.03}$ & \scriptsize $-4.12$ & $\scriptscriptstyle{\pm 0.03}$ & \scriptsize $-4.29$ & $\scriptscriptstyle{\pm 0.03}$  \\ \midrule 
\small Two Moons & \scriptsize $\underline{0.002}$ & $\scriptscriptstyle{\pm 0.001}$ & \scriptsize $0.023$ & $\scriptscriptstyle{\pm 0.002}$ & \scriptsize $0.008$ & $\scriptscriptstyle{\pm 0.002}$ & \scriptsize $1.40$ & $\scriptscriptstyle{\pm 0.52}$ & \scriptsize $\underline{-0.52}$ & $\scriptscriptstyle{\pm 0.10}$ & \scriptsize $1.36$ & $\scriptscriptstyle{\pm 0.51}$ & \scriptsize $-1.02$ & $\scriptscriptstyle{\pm 0.01}$ & \scriptsize $\underline{-0.36}$ & $\scriptscriptstyle{\pm 0.01}$ & \scriptsize $-0.95$ & $\scriptscriptstyle{\pm 0.01}$  \\ \midrule 
\small Trajectories & \scriptsize $\underline{0.008}$ & $\scriptscriptstyle{\pm 0.002}$ & \scriptsize $0.016$ & $\scriptscriptstyle{\pm 0.001}$ & \textcolor{red}{\scriptsize $0.743$} & \textcolor{red}{$\scriptscriptstyle{\pm 0.005}$} & \scriptsize $\underline{1.03}$ & $\scriptscriptstyle{\pm 0.22}$ & \scriptsize $1.24$ & $\scriptscriptstyle{\pm 0.24}$ & \textcolor{red}{\scriptsize $9.30$} & \textcolor{red}{$\scriptscriptstyle{\pm 1.30}$} & \scriptsize $\underline{29.32}$ & $\scriptscriptstyle{\pm 0.02}$ & \scriptsize $26.09$ & $\scriptscriptstyle{\pm 0.02}$ & \textcolor{red}{\scriptsize $-215.23$} & \textcolor{red}{$\scriptscriptstyle{\pm 17.6}$}  \\ \bottomrule[1pt]
\end{tabular} 
\caption{Baseline Comparison -- marked in red are cases with notably poor performance; best values are underlined.}
\label{tab: baselines}
\end{table*}
\section{Experiments}
\label{sec:experiments}
We compare our approach against two baselines from the literature (VC~\cite{nagler2016evading} and MPW~\cite{vincent2002manifold}) in four scenarios, using three metrics. Additionally, we carry out an ablation study on our proposed method ROME.

For the hyperparameters pertaining to the clustering within ROME (see Section~\ref{sec:clustering}), we found empirically that stable results can be obtained using 199 possible clusterings, 100 for DBSCAN (Equation~\eqref{eq:DBSCAN})
\begin{equation*}\begin{aligned}
    \bm{\varepsilon} = & \left\{ \min R_N + \left(\frac{\alpha}{99}\right)^2\left( \max (R_N\backslash\{\infty\}) - \min R_N \right)\vert \right. \\ & \left. \vphantom{\frac{\alpha}{99}} \, \, \, \, \alpha \in \{0, \hdots, 99\} \right\}
\end{aligned}\end{equation*}
combined with 99 for $\xi$-clustering (Equation~\eqref{eq:xi})
\begin{equation*}
    \bm{\xi} = \left\{\frac{\beta}{100} \, \vert \, \beta \in \{1, \hdots, 99\} \right\},
\end{equation*}
as well as using $k_{\min} = 5$, $k_{\max} = 20$, and $\alpha_k = 400$ for calculating $k_{\bm{c}}$ (Equation~\eqref{eq:kC} and Appendix C).

\subsection{Distributions}
In order to evaluate different aspects of a density estimation method $f$, we used a number of different distributions. 

\begin{itemize}
    \item Three two-dimensional synthetic distributions (Figure~\ref{fig:2Ddistributions}) were used to test the estimation of distributions with multiple clusters, which might be highly correlated (Aniso) or of varying densities (Varied), or express non-normal distributions (Two Moons).
    \item A multivariate, 24-dimensional, and highly correlated distribution generated from a subset of the Forking Paths dataset \cite{liang2020garden} (Figure~\ref{fig:bimodal}). The 24 dimensions correspond to the $x$ and $y$ positions of a human pedestrian across $12$ timesteps.
    Based on 6 original trajectories ($\bm{x}^*_i \in \mathbb{R}^{12 \times 2}$), we defined the underlying distribution $p$ in such a way, that one could calculate a sample $\bm{x} \sim p$ with:
    \begin{equation*}
        \bm{x} = s \bm{x}^*_i \bm{R}_{\theta}^T + \bm{L}\bm{n} \, , \, \text{with } i \sim \mathcal{U}\{1,6\} \,. 
    \end{equation*}
    Here, $\bm{R}_\theta \in \mathbb{R}^{2\times 2}$ is a rotation matrix rotating $\bm{x}^*_i$ by $\theta \sim \mathcal{N}(0, \frac{\pi}{180})$, while $s \sim \mathcal{N}(1, 0.03)$ is a scaling factor. $\bm{n} = \mathcal{N}(\bm{0}, 0.03\bm{I}) \in \mathbb{R}^{12 \times 2}$ is additional noise added on all dimensions using $\bm{L} \in \mathbb{R}^{12\times 12}$, a lower triangular matrix that only contains ones.
\end{itemize}
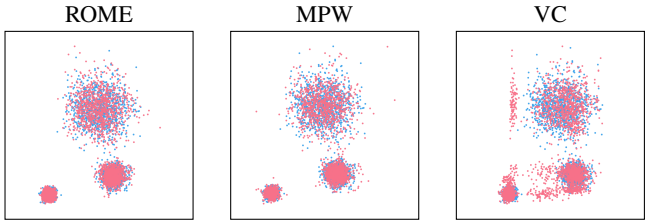
\begin{figure}[!t]
    \centering
    \vspace{-4mm}
    \begin{tikzpicture}
        \begin{scope}[xshift = 0cm]
            \node[above right, inner sep = 0] at (0,0) {\includegraphics[width = 2.5cm]{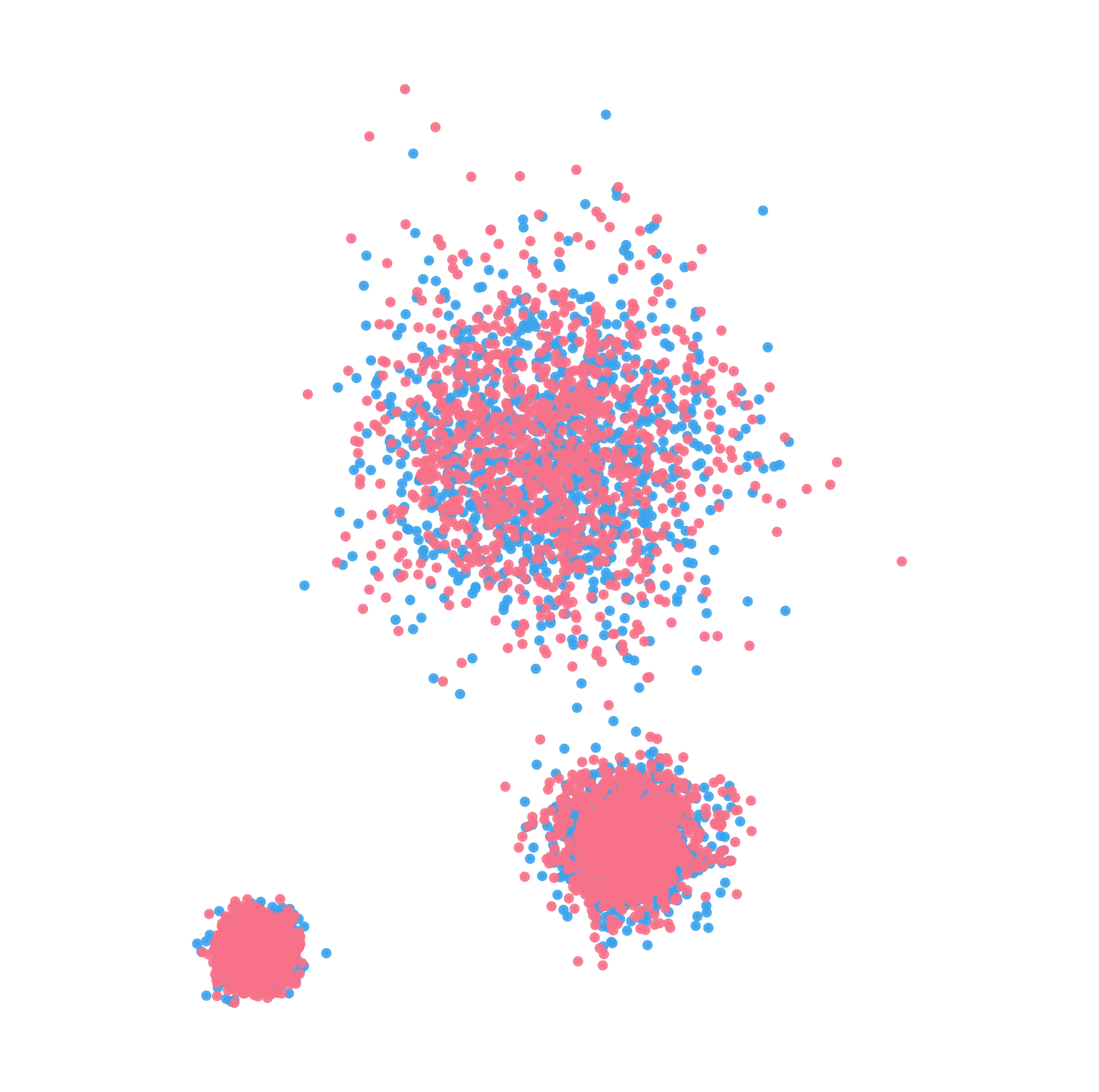}};
            \draw[black] (0,0) rectangle (2.5,2.5);
            \node[above, black, font = \small] at (1.25,2.5) {ROME};
        \end{scope}
        \begin{scope}[xshift = 3cm]
            \node[above right, inner sep = 0] at (0,0) {\includegraphics[width = 2.5cm]{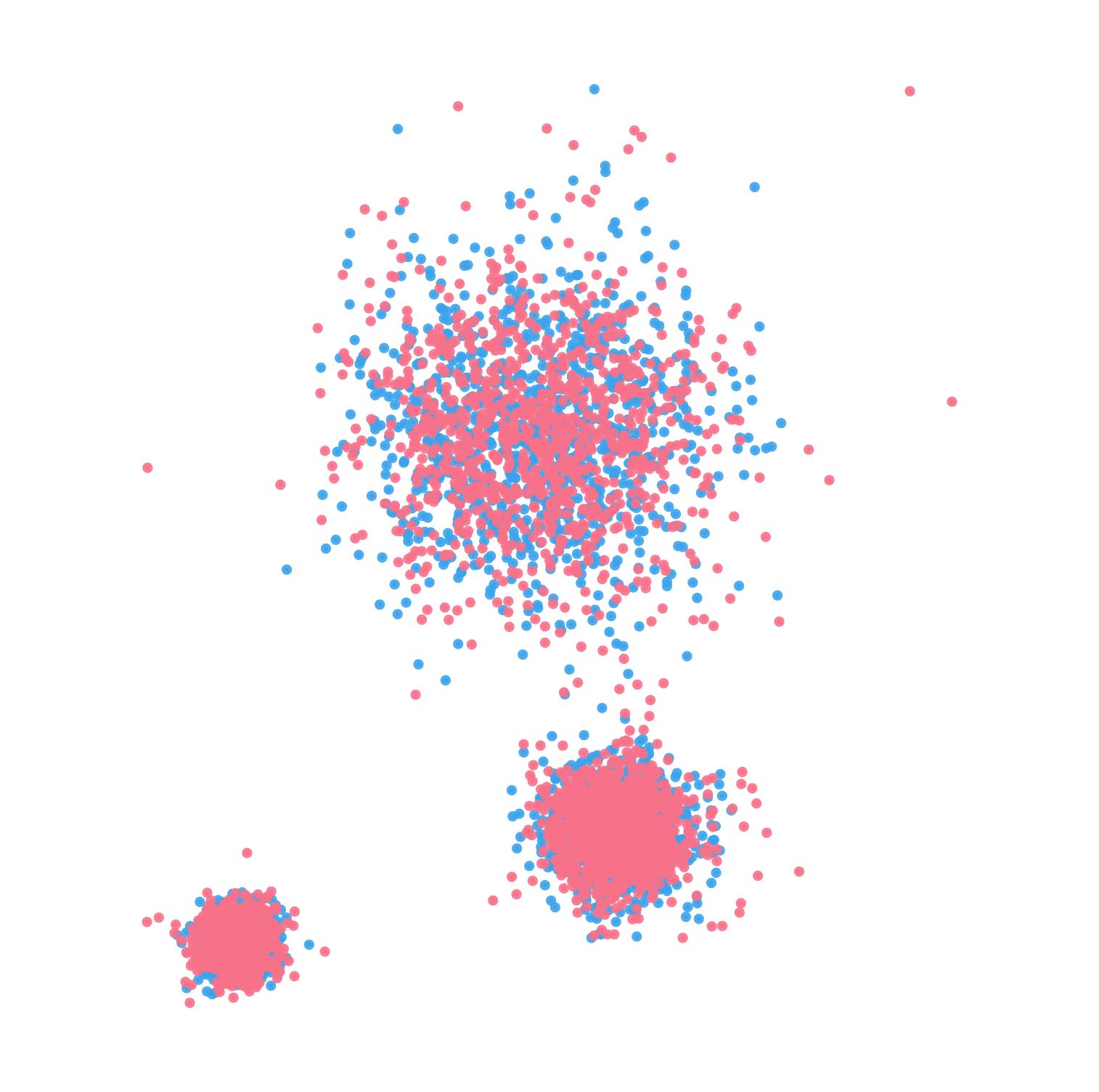}};
            \draw[black] (0,0) rectangle (2.5,2.5);
            \node[above, black, font = \small] at (1.25,2.5) {MPW};
        \end{scope}
        \begin{scope}[xshift = 6cm]
            \node[above right, inner sep = 0] at (0,0) {\includegraphics[width = 2.5cm]{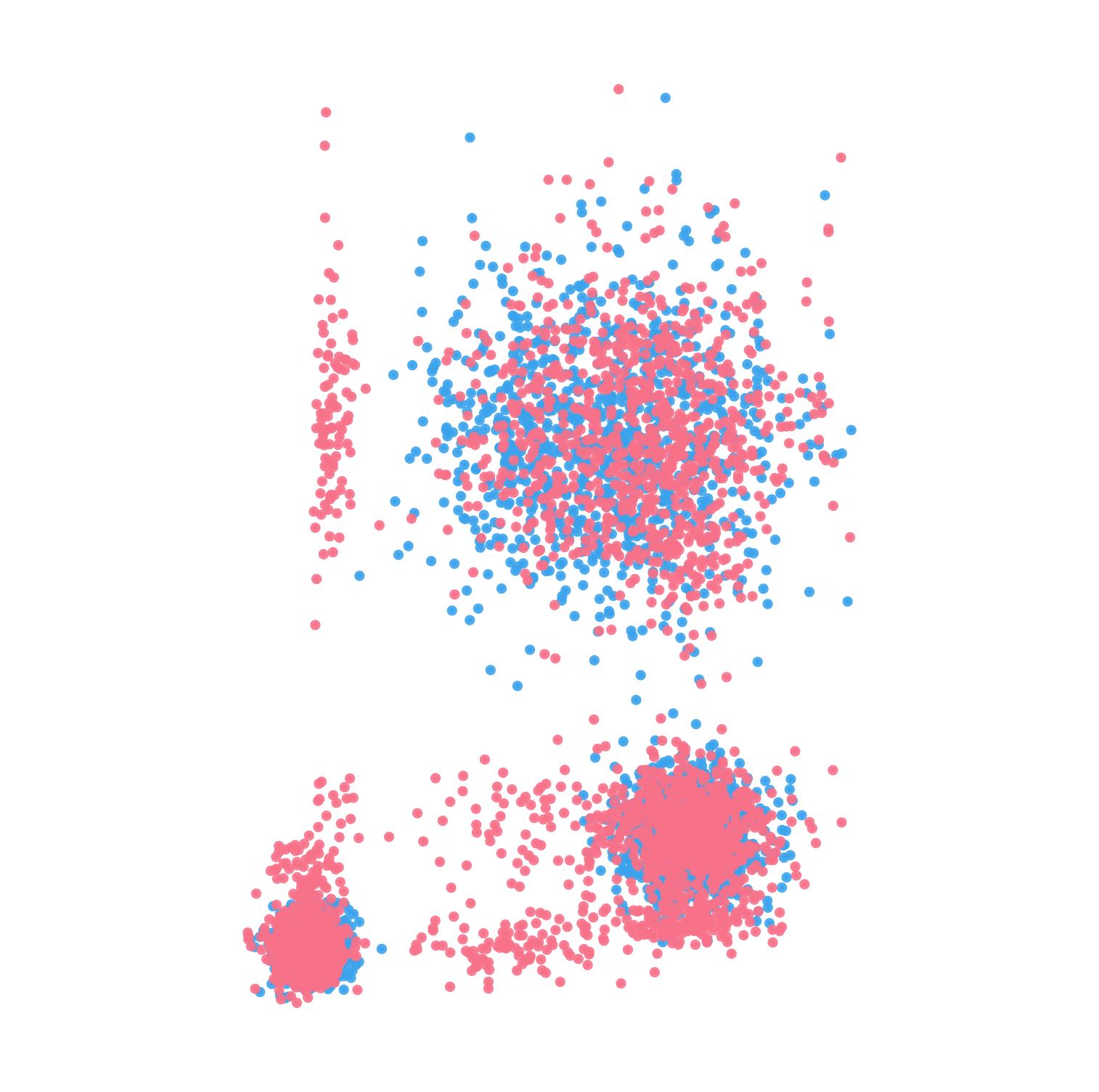}};
            \draw[black] (0,0) rectangle (2.5,2.5);
            \node[above, black, font = \small] at (1.25,2.5) {VC};
        \end{scope}
    \end{tikzpicture}
    \vspace{-4mm}
    \caption{Samples obtained with ROME, MPW and VC (pink) contrasted with samples from $p$ (blue); Varied. 
    }
    \vspace{-2mm}
    \label{fig:baselinePlots}
\end{figure}
Further tests on uni-modal problems can be found in Appendix E. 

\subsection{Evaluation and Metrics}

When estimating density $\widehat{p}$, since we cannot query the distribution $p$ underlying the samples $\bm{X}$, we require metrics that can provide insights purely based on those samples.
To this end we use the following three metrics to quantify how well a given density estimator $f$ can avoid both over-fitting and over-smoothing.
\begin{itemize}
    \item To test for over-fitting, we first sample two different datasets $\bm{X}_1$ and $\bm{X}_2$ ($N$ samples each) from $p$ ($\bm{X}_1, \bm{X}_2 \sim p$). We then use the estimator $f$ to create two queryable distributions $\widehat{p}_{1} = f(\bm{X}_1)$ and $\widehat{p}_{2} = f(\bm{X}_2)$. If those distributions $\widehat{p}_{1}$ and $\widehat{p}_{2}$ are similar, it would mean the tested estimator does not over-fit; we measure this similarity using the Jensen-Shannon divergence~\cite{lin1991divergence}:
    \begin{equation*}
    \begin{aligned}
        D_\text{JS}(\widehat{p}_{1}\Vert \widehat{p}_{2}) &= \frac{1}{2N \ln(2)} \sum\limits_{\bm{x} \in \bm{X}_1 \cup \bm{X}_2} h_1(\bm{x})  + h_2(\bm{x}) \\
        h_i(\bm{x}) &= \frac{\widehat{p}_{i} (\bm{x})}{\widehat{p}_{1} (\bm{x}) + \widehat{p}_{2} (\bm{x})} \ln \left( \frac{2\widehat{p}_{i} (\bm{x})}{\widehat{p}_{1} (\bm{x}) + \widehat{p}_{2} (\bm{x})} \right)
    \end{aligned}
    \end{equation*}
    This metric, however, is not able to account for systematic biases that could be present in the estimator $f$. Comparisons of $\widehat{p}_1$ with $p$ -- if the dataset allows -- can be found in Appendix D. 

    \item To test the goodness-of-fit of the estimated density, we first generate a third set of samples $\widehat{\bm{X}}_1 \sim \widehat{p}_{1}$ with $N$ samples. We then use the Wasserstein distance $W$~\cite{villani2009optimal} on the data to calculate the indicator $\widehat{W}$:
    \begin{equation*}
        \widehat{W} = \frac{W(\bm{X}_1, \widehat{\bm{X}}_1) - W(\bm{X}_1, \bm{X}_2)}{W(\bm{X}_1, \bm{X}_2)}
    \end{equation*}
    Here, $\widehat{W} > 0$ indicates over-smoothing or misplaced modes, while $-1 \leq \widehat{W} < 0$ indicates over-fitting.

    \item Not every density estimator $f$ has the ability to generate the samples $\widehat{\bm{X}}_1$. Consequently, we need to test for goodness-of-fit without relying on $\widehat{\bm{X}}_1$. Therefore, we use the average log-likelihood 
    \begin{equation*}
        \widehat{L} = \frac{1}{N} \sum\limits_{\bm{x} \in \bm{X}_2} \ln\left(\widehat{p}_1 (\bm{x}) \right) \, ,
    \end{equation*}
    which would be maximized only for $p = \widehat{p}_1$ as long as $\bm{X}_2$ is truly representative of $p$ (see Gibbs' inequality~\cite{kvalseth1997generalized}). However, using this metric might be meaningless if $\widehat{p}_1$ is not normalized, as the presence of the unknown scaling factor makes the $\widehat{L}$ values of two estimators incomparable, and cannot discriminate between over-fitting and over-smoothing.
\end{itemize}

For each candidate method $f$, we used $N = 3000$, and every metric calculation was repeated $100$ times to take into account the inherent randomness of sampling from the distributions, with the standard deviation in the tables being reported with $\scriptscriptstyle{\pm}$.

\subsection{Ablations}
To better understand the performance of our approach, we investigated variations in four key aspects of ROME:
\begin{itemize}
    \item \textit{Clustering approach}.  First, we replaced the silhouette score (see Section~\ref{sec:clustering}) with density based cluster validation (DBCV)~\cite{moulavi2014density} when selecting the optimal clustering out of the 199 possibilities. Furthermore, we investigated the approach of no clustering ($\bm{C} = \{\{1, \hdots, N \}\}$) .
    \item \textit{Decorrelation vs No decorrelation}. We investigated the effect of removing rotation by setting $\bm{R}_C = \bm{I}$. 
    \item \textit{Normalization vs No normalization}. We studied the sensitivity of our approach to normalization by setting $\widetilde{\bm{\Sigma}}_{C} = \bm{I}$. 
    \item \textit{Downstream density estimator}. We  replaced $f_{\text{KDE}}$ with two other candidate methods. First, we used a single-component Gaussian mixture model $f_{\text{GMM}}$
    \begin{equation*}
        f_{\text{GMM}}(\bm{X})(\bm{x}) = \mathcal{N}\left(\bm{x} \vert \bm{\widehat{\mu}}_{\bm{X}}, \bm{\widehat{\Sigma}}_{\bm{X}}\right)
    \end{equation*}
    fitted to the observed mean $\bm{\widehat{\mu}}$ and covariance matrix $\bm{\widehat{\Sigma}}$ of a dataset $\bm{X}$. 
    Second, we used a $k$-nearest neighbor approach $f_{\text{kNN}}$~\cite{loftsgaarden1965nonparametric}
    \begin{equation*}
        f_{\text{kNN}}(\bm{X})(\bm{x}) = \frac{k}{N \; \mathcal{V}_M \; \underset{\bm{\widehat{x}} \in \bm{X}}{{\min}_{k}}  \left\Vert \bm{x} - \bm{\widehat{x}} \right\Vert^M}
    \end{equation*}
    where $\mathcal{V}_M$ is the volume of the $M$-dimensional unit hypersphere. We used the rule-of-thumb $k = \lfloor \sqrt{N} \rfloor$. However, this estimator cannot generate samples.
\end{itemize}

While those four factors would theoretically lead to 24 estimators, $f_{\text{KDE}}$ as well as $f_{\text{kNN}}$ being invariant against rotation and $f_{\text{GMM}}$ being invariant against any linear transformation means that only $14$ of ROME's ablations are actually unique.

\section{Results}

\subsection{Baseline Comparison}

\begin{table}[t!]
\centering
\begin{tabularx}{\linewidth}{X | Z | Z | Z | Z}
\toprule[1pt] 
& \multicolumn{2}{c|}{Norm.} & \multicolumn{1}{c|}{No norm.} & \multicolumn{1}{c}{\multirow{2}{*}{$f_{\text{GMM}}$}} \\
Cluster. & \multicolumn{1}{c|}{Decorr.} & \multicolumn{2}{c|}{No decorr.} & \\ \midrule[1pt]
\small Silhouette & \cellcolor{gray!25}\scriptsize $-0.13\scriptscriptstyle{\pm 0.20}$ & \scriptsize $-0.13\scriptscriptstyle{\pm 0.20}$ & \scriptsize $-0.13\scriptscriptstyle{\pm 0.19}$ & \scriptsize $-0.14\scriptscriptstyle{\pm 0.20}$ \\ \midrule 
\small DBCV & \scriptsize $-0.09\scriptscriptstyle{\pm 0.23}$ & \scriptsize $-0.09\scriptscriptstyle{\pm 0.23}$ & \scriptsize $-0.07\scriptscriptstyle{\pm 0.23}$ & \scriptsize ${-0.03}\scriptscriptstyle{\pm 0.24}$ \\ \midrule 
\small No clus. & \textcolor{red}{\scriptsize $\hphantom{-}2.28\scriptscriptstyle{\pm 0.72}$}& \textcolor{red}{\scriptsize $\hphantom{-}2.26\scriptscriptstyle{\pm 0.72}$} & \scriptsize $-0.17\scriptscriptstyle{\pm 0.26}$ & \textcolor{red}{\scriptsize $\:\, 10.53\scriptscriptstyle{\pm 2.54}$} \\ \bottomrule[1pt]
\end{tabularx} 
\caption{Ablations ($\widehat{W} \rightarrow 0$, Varied): Clustering is essential to prevent over-smoothing. ROME highlighted in gray. Note that the differences between Silhouette and DBCV are not statistically significant.}
\label{tab: WS_varied}
\end{table}
\begin{table}[t!]
\centering
\begin{tabularx}{\linewidth}{X | Z | Z | Z }
\toprule[1pt] 
& \multicolumn{2}{c|}{Norm.} & \multicolumn{1}{c}{No norm.} 
\\
Cluster. & \multicolumn{1}{c|}{Decorr.} & \multicolumn{2}{c}{No decorr.} 
\\ \midrule[1pt]
\small Silhouette & {\cellcolor{gray!25}\scriptsize $-2.53\scriptscriptstyle{\pm 0.02}$} & {\scriptsize $-2.70\scriptscriptstyle{\pm 0.01}$} & {\scriptsize $-2.79\scriptscriptstyle{\pm 0.01}$} 
\\ \midrule 
\small DBCV & {\scriptsize $-2.56\scriptscriptstyle{\pm 0.02}$} & {\scriptsize $-2.69\scriptscriptstyle{\pm 0.01}$} & {\scriptsize $-2.83\scriptscriptstyle{\pm 0.02}$} 
\\ \bottomrule[1pt]
\end{tabularx}
\caption{Ablations ($\widehat{L} \uparrow$, Aniso): When clustering, decorrelation and normalization improve results for distributions with high intra-mode correlation. ROME highlighted in gray.}
\label{tab: log_aniso}
\end{table}
We found that ROME avoids the major pitfalls displayed by the two baseline methods on the four tested distributions (Table~\ref{tab: baselines}). Out of the two baseline methods, the manifold Parzen windows (MPW) approach has a stronger tendency to over-fit in the case of the two-dimensional distributions compared to ROME, as quantified by lower $D_\text{JS}$ values achieved by ROME.
MPW does, however, achieve a better log-likelihood for the Two Moons distribution compared to ROME. This could be due to the locally adaptive non-Gaussian distributions being less susceptible to over-smoothing than our approach of using a single isotropic kernel for each cluster if such clusters are highly non-normal. 
Lastly, in the case of the pedestrian trajectories distribution, ROME once more achieves better performance than MPW, with MPW performing worse both in terms of $D_\text{JS}$ and $\widehat{L}$.

Meanwhile, the vine copulas (VC) approach tends to over-smooth the estimated densities (large positive $\widehat{W}$ values), and even struggles with capturing the different modes (see Figure \ref{fig:baselinePlots}).
This is likely because VC uses KDE with Silverman's rule of thumb, which is known to lead to over-smoothing in the case of multi-modal distributions~\cite{heidenreich2013bandwidth}.
Furthermore, on the pedestrian trajectories distribution, we observed both high $D_\text{JS}$ and $\widehat{W}$ values, indicating that VC is unable to estimate the underlying density; this is also indicated by the poor log-likelihood estimates.

Overall, while the baselines were able to achieve better performance in selected cases (e.g., MPW better than ROME in terms of $\widehat{W}$ and $\widehat{L}$ on the Two Moons distribution), they have their apparent weaknesses. Specifically, MPW achieves poor results for most metrics in the case of varying densities within the modes (Varied), while Vine Copulas obtain the worst performance across all three metrics in the case of the multivariate trajectory distributions.
ROME, in contrast, achieved high performance across all the test cases.
\begin{table}[t!]
\centering
\begin{tabularx}{\linewidth}{X | Z | Z | Z | Z} 
\toprule[1pt] 
& \multicolumn{2}{c|}{Norm.} & \multicolumn{1}{c|}{No norm.} & \multirow{2}{*}{$f_{\text{GMM}}$} \\
Cluster. & \multicolumn{1}{c|}{Decorr.} & \multicolumn{2}{c|}{No decorr.} & \\ \midrule[1pt]
\small Silhouette & {\cellcolor{gray!25}\scriptsize ${1.40}\scriptscriptstyle{\pm 0.52}$} & {\scriptsize $1.41\scriptscriptstyle{\pm 0.52}$} & \textcolor{red}{\scriptsize $3.65\scriptscriptstyle{\pm 0.99}$} & \textcolor{red}{\scriptsize $4.37\scriptscriptstyle{\pm 1.14}$}  \\ \midrule 
\small DBCV & {\scriptsize $1.59\scriptscriptstyle{\pm 0.56}$} & {\scriptsize $1.59\scriptscriptstyle{\pm 0.56}$} & \textcolor{red}{\scriptsize $4.24\scriptscriptstyle{\pm 1.13}$} & \textcolor{red}{\scriptsize $4.77\scriptscriptstyle{\pm 1.23}$}  \\ \midrule 
\small No clus. & {\scriptsize $1.82\scriptscriptstyle{\pm 0.60}$} & {\scriptsize $1.84\scriptscriptstyle{\pm 0.60}$} & \textcolor{red}{\scriptsize $3.17\scriptscriptstyle{\pm 0.89}$} & \textcolor{red}{\scriptsize $5.29\scriptscriptstyle{\pm 1.34}$}  \\ \bottomrule[1pt]
\end{tabularx} 
\caption{Ablations ($\widehat{W} \rightarrow 0$, Two Moons): Excluding normalization or using $f_{\text{GMM}}$ as the downstream estimator is not robust against non-normal distributions. ROME highlighted in gray.}
\label{tab: WS_twoMoons}
\end{table}
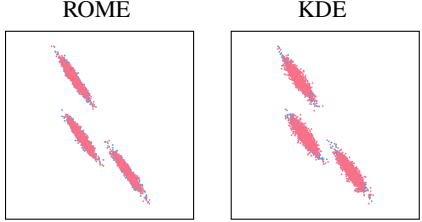
\begin{figure}[t!]
    \centering
    \vspace{-4mm}
    \begin{tikzpicture}
        \begin{scope}[xshift = 0cm]
            \node[above right, inner sep = 0] at (0,0) {\includegraphics[width = 2.5cm]{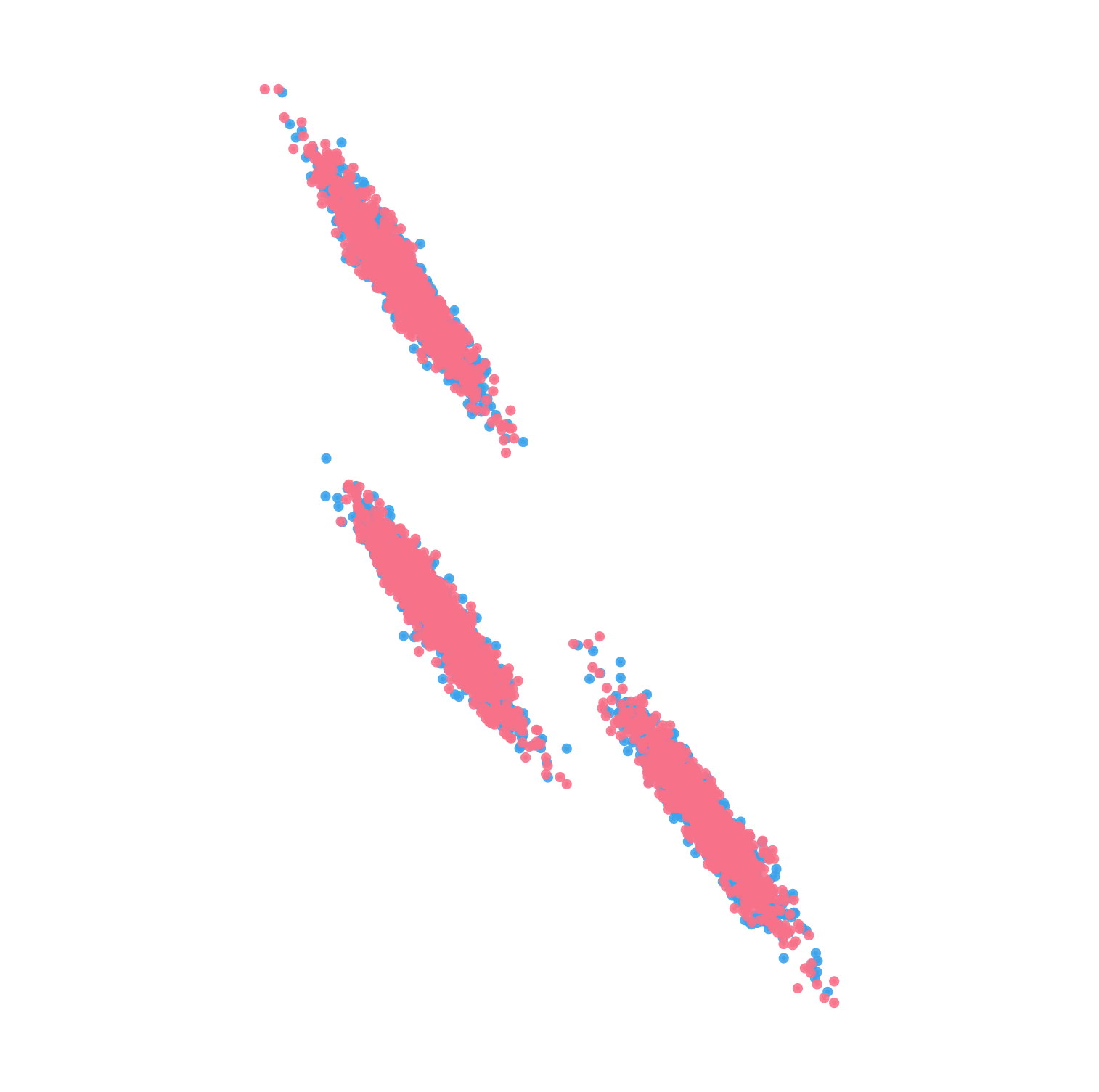}};
            \draw[black] (0,0) rectangle (2.5,2.5);
            \node[above, black, font = \small] at (1.25,2.5) {ROME \vphantom{$f_{\text{GMM}}$}};
        \end{scope}
        \begin{scope}[xshift = 3cm]
            \node[above right, inner sep = 0] at (0,0) {\includegraphics[width = 2.5cm]{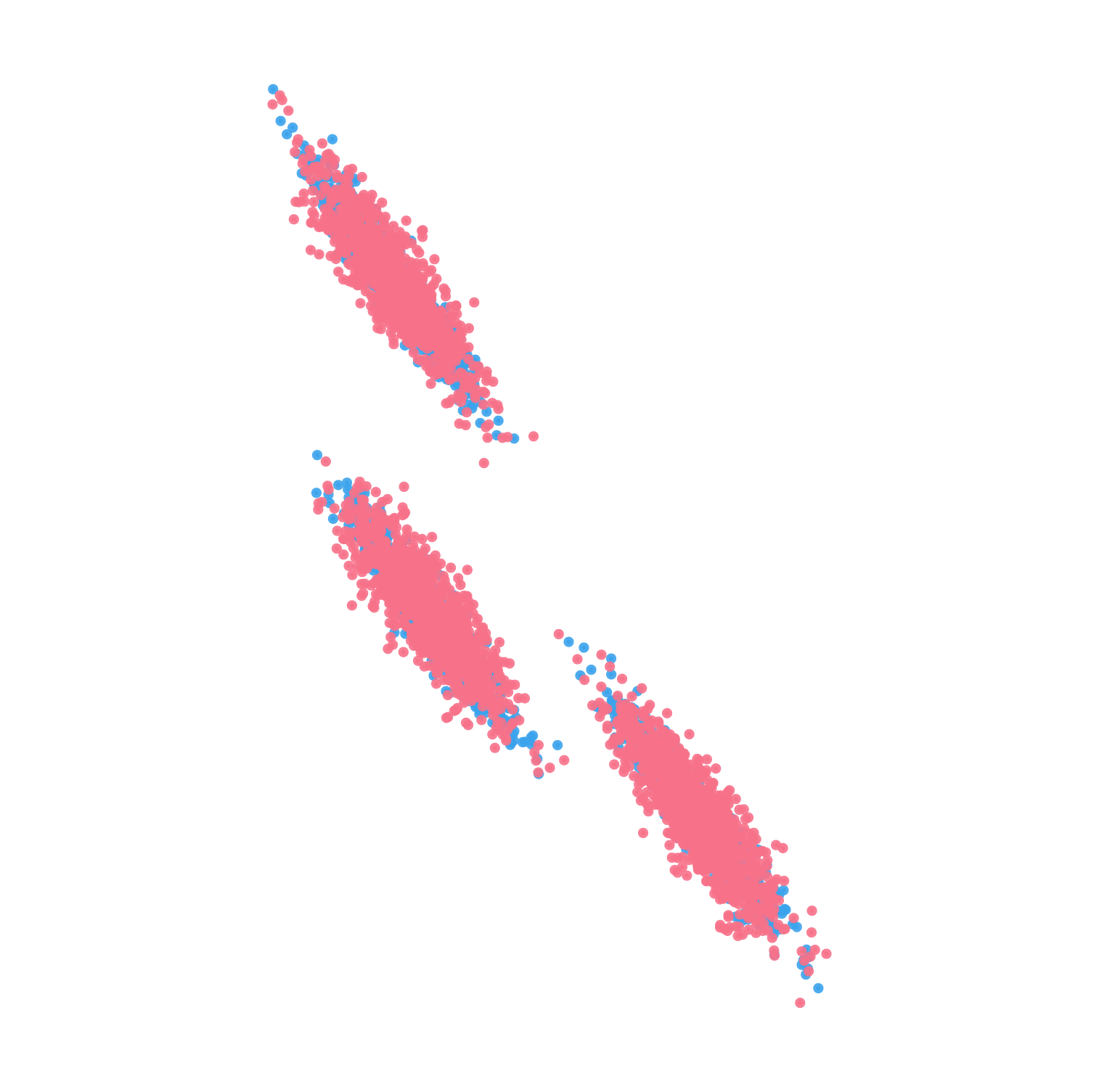}};
            \draw[black] (0,0) rectangle (2.5,2.5);
            \node[above, black, font = \small] at (1.25,2.5) {KDE \vphantom{$f_{\text{GMM}}$}};
        \end{scope}
    \end{tikzpicture}
    \vspace{-2mm}
    \caption{Samples generated by ROME and KDE -- equivalent to ROME without clustering, decorrelation and normalization -- (in pink) contrasted with samples from $p$ (in blue); Aniso. Note that the samples by KDE are more spread out, indicating over-smoothing.}
    \vspace{-2mm}
    \label{fig:KDEplot}
\end{figure}
\begin{table*}[!t]
\centering
\begin{tabularx}{\textwidth}{X | J  J | J  J | J  J | J} 
\toprule[1pt] 
& \multicolumn{4}{c|}{Normalization} & \multicolumn{2}{c|}{No normalization} & \multirow{3}{*}{$f_{\text{GMM}}$} \\
& \multicolumn{2}{c|}{Decorrelation} & \multicolumn{4}{c|}{No decorrelation} & \\ 
Clustering & $f_{\text{KDE}}$ & $f_{\text{kNN}}$ & $f_{\text{KDE}}$ & $f_{\text{kNN}}$ & $f_{\text{KDE}}$ & $f_{\text{kNN}}$ & \\
\midrule[1pt]
\small Silhouette & {\cellcolor{gray!25}\scriptsize $0.084\scriptscriptstyle{\pm 0.016}$} & \textcolor{red}{\scriptsize $1.045\scriptscriptstyle{\pm 0.064}$} & {\scriptsize $0.777\scriptscriptstyle{\pm 0.116}$} & \textcolor{red}{\scriptsize $1.808\scriptscriptstyle{\pm 0.112}$} & {\scriptsize $0.015\scriptscriptstyle{\pm 0.011}$} & \textcolor{red}{\scriptsize $1.887\scriptscriptstyle{\pm 0.118}$} & {\scriptsize $0.032\scriptscriptstyle{\pm 0.007}$}  \\ \midrule 
\small DBCV & {\scriptsize $0.090\scriptscriptstyle{\pm 0.015}$} & \textcolor{red}{\scriptsize $1.119\scriptscriptstyle{\pm 0.073}$} & {\scriptsize $0.897\scriptscriptstyle{\pm 0.154}$} & \textcolor{red}{\scriptsize $1.937\scriptscriptstyle{\pm 0.109}$} & {\scriptsize $0.017\scriptscriptstyle{\pm 0.012}$} & \textcolor{red}{\scriptsize $1.934\scriptscriptstyle{\pm 0.116}$} & {\scriptsize $0.043\scriptscriptstyle{\pm 0.010}$}  \\ \midrule 
\small No clusters & {\scriptsize $0.009\scriptscriptstyle{\pm 0.004}$} & \textcolor{red}{\scriptsize $0.453\scriptscriptstyle{\pm 0.051}$} & {\scriptsize $0.015\scriptscriptstyle{\pm 0.012}$} & \textcolor{red}{\scriptsize $1.044\scriptscriptstyle{\pm 0.104}$} & {\scriptsize $0.005\scriptscriptstyle{\pm 0.003}$} & \textcolor{red}{\scriptsize $1.478\scriptscriptstyle{\pm 0.132}$} & {\scriptsize $0.017\scriptscriptstyle{\pm 0.011}$}  \\ \bottomrule[1pt]
\end{tabularx} 
\caption{Ablations ($D_\text{JS} \downarrow_{0}^{1}$, Trajectories; values are multiplied by $10$ for easier comprehension): Using $f_{\text{kNN}}$ as the downstream estimator tends to lead to over-fitting. ROME highlighted in gray.}
\label{tab: JSD_traj}
\end{table*}
\subsection{Ablation Studies}
When it comes to the choice of the clustering method, our experiments show no clear advantage for using either the silhouette score or DBCV. 
But as the silhouette score is computationally more efficient than DBCV, it is the preferred method.
However, using clustering is essential, as otherwise there is a risk of over-smoothing, such as in the case of multi-modal distributions with varying densities in each mode (Table~\ref{tab: WS_varied}).

Testing variants of ROME on the Aniso distribution (Table~\ref{tab: log_aniso}) demonstrated not only the need for decorrelation, through the use of rotation, but also normalization in the case of distributions with highly correlated features. 
There, using either of the two clustering methods in combination with normalization and decorrelation (our full proposed method) is better than the two alternatives of omitting only decorrelation or both decorrelation and normalization. 
In the case of clustering with the silhouette score, the full method is significantly more likely to reproduce the underlying distribution $p$ by a factor of $1.19$ (with a statistical significance of $10 ^{-50}$, see Appendix B) as opposed to omitting only decorrelation, and by $1.30$ compared to omitting both decorrelation and normalization. 
Similar trends can be seen when clustering based on DBCV, with the full method being more likely to reproduce $p$ by a factor of $1.14$ and $1.31$ respectively. 
Results on the Aniso distribution further show that KDE on its own is not able to achieve the same results as ROME, but rather it has a tendency to over-smooth (Figure~\ref{fig:KDEplot}).
Additionally, the ablation with and without normalization on the Two Moons distribution (Table~\ref{tab: WS_twoMoons}) showed that normalization is necessary to avoid over-smoothing on non-normal distributions.

\begin{figure}[t!]
    \centering
    \vspace{-4mm}
    \begin{tikzpicture}
        \begin{scope}[xshift = 0cm]
            \node[above right, inner sep = 0] at (0,0) {\includegraphics[width = 2.5cm]{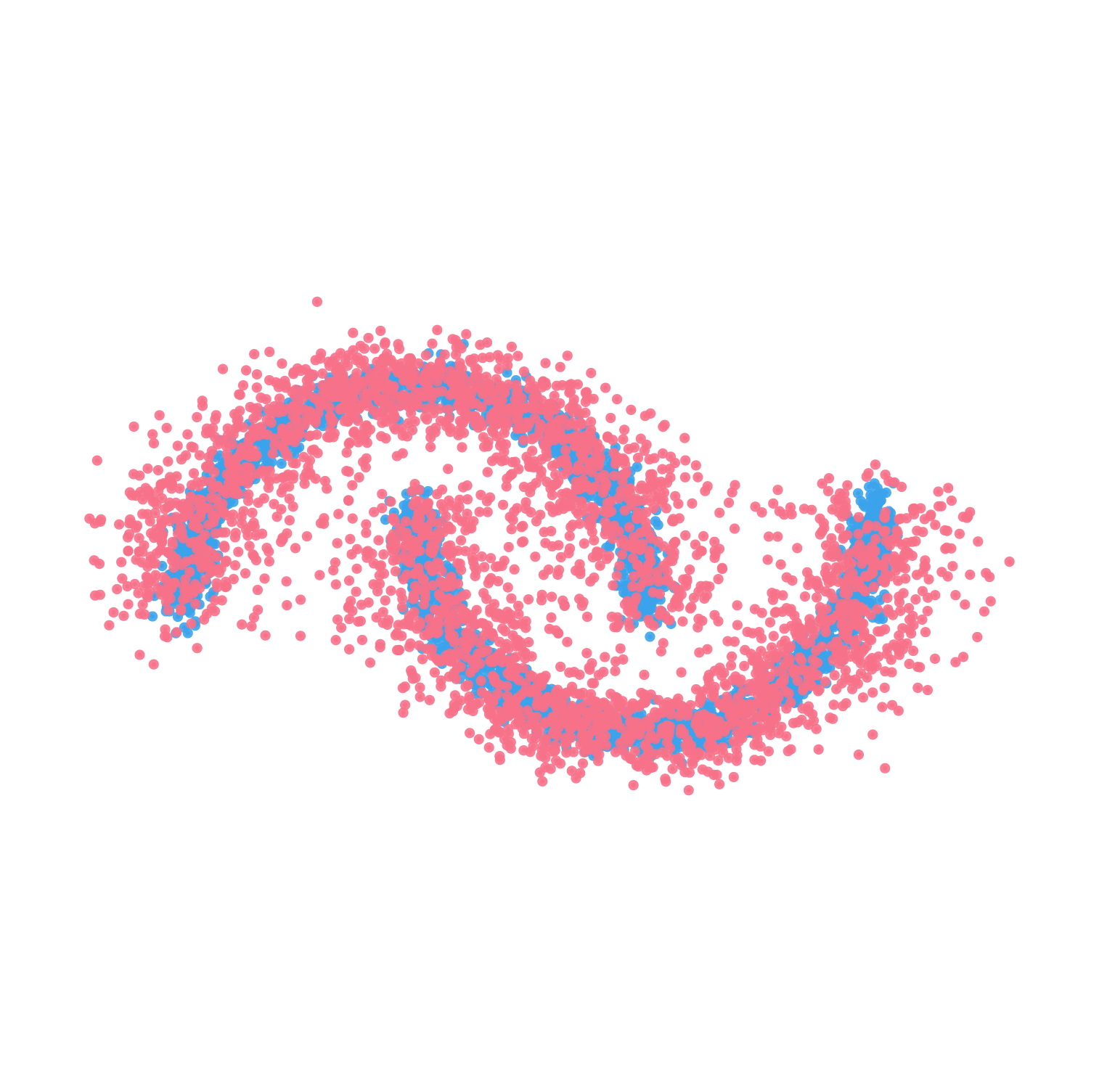}};
            \draw[black] (0,0) rectangle (2.5,2.5);
            \node[above, black, font = \small] at (1.25,2.5) {ROME \vphantom{$f_{\text{GMM}}$}};
        \end{scope}
        \begin{scope}[xshift = 3cm]
            \node[above right, inner sep = 0] at (0,0) {\includegraphics[width = 2.5cm]{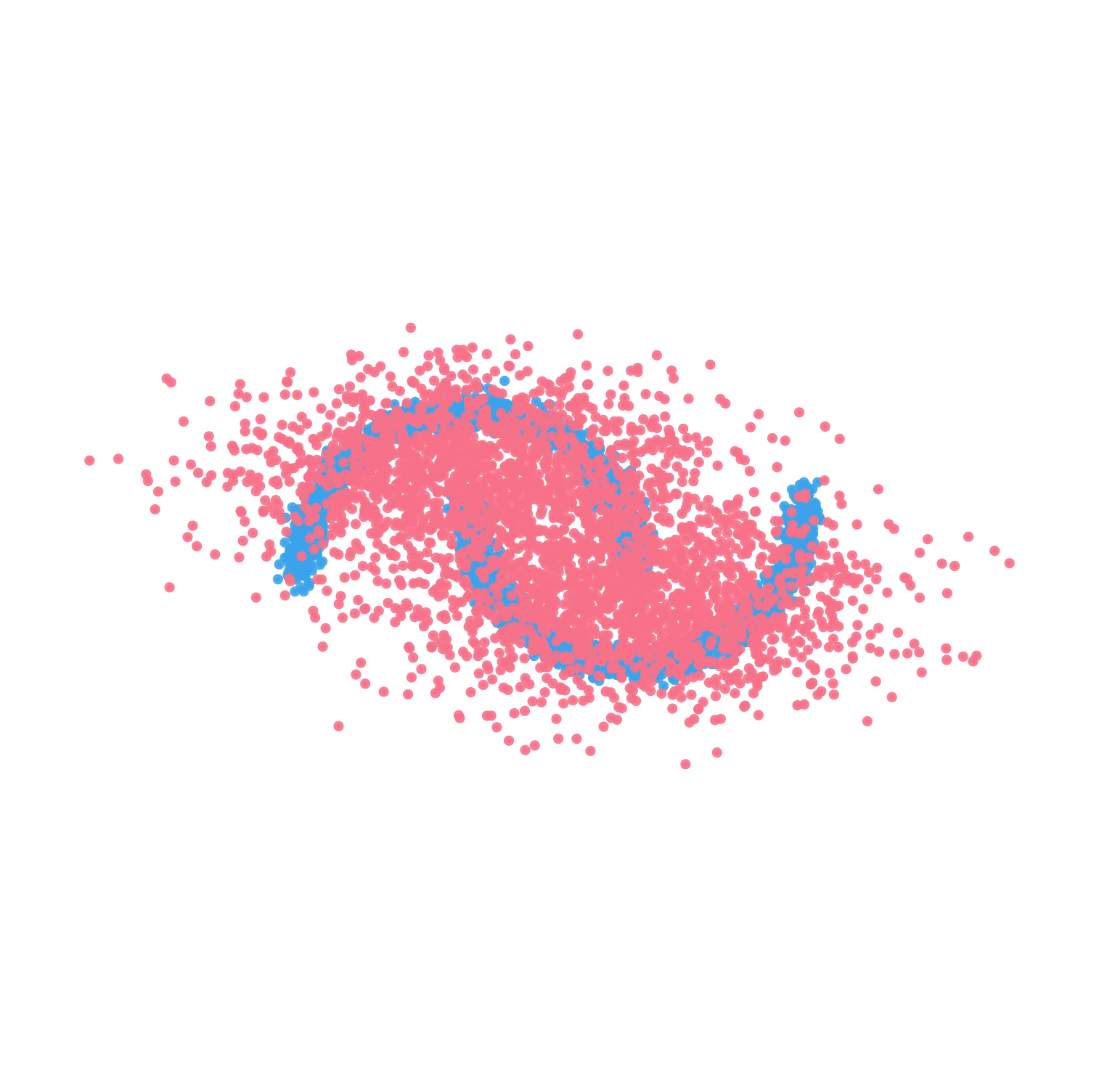}};
            \draw[black] (0,0) rectangle (2.5,2.5);
            \node[above, black, font = \small] at (1.25,2.5) {ROME with $f_{\text{GMM}}$};
        \end{scope}
    \end{tikzpicture}
    \vspace{-2mm}
    \caption{Samples generated by ROME, and ROME with $f_{\text{GMM}}$ as the downstream estimator (in pink) contrasted with samples from $p$ (in blue); Two Moons. Note that the samples from $f_{\text{GMM}}$ are more spread out, which clearly displays over-smoothing.}
    \vspace{-2mm}
    \label{fig:GMMplot}
\end{figure}
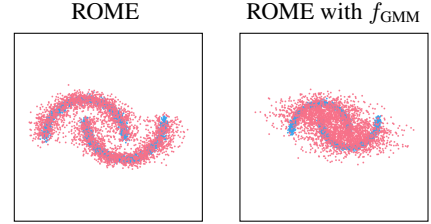

Lastly, investigating the effect of different downstream density estimators, we found that using ROME with $f_{\text{kNN}}$ instead of $f_{\text{KDE}}$ leads to over-fitting (highest $D_\text{JS}$ values in Table~\ref{tab: JSD_traj}). 
Meanwhile, ROME with $f_{\text{GMM}}$ tends to over-smooth the estimated density in cases where the underlying distribution is not Gaussian (high $\widehat{W}$ in Table~\ref{tab: WS_twoMoons}).
The over-smoothing caused by $f_{\text{GMM}}$ is further visualised in Figure~\ref{fig:GMMplot}.

In conclusion, our ablation studies confirmed that using $f_{\text{KDE}}$ in combination with data clustering, normalization and decorrelation provides the most reliable density estimation for different types of distributions.
\section{Conclusion}
In our comparison against two established and sophisticated density estimators, we observed that ROME achieved consistently good results across all tests, while manifold Parzen windows (MPW) and vine copulas (VC) were susceptible to over-fitting and over-smoothing. 
For example, while MPW is superior at capturing non-normal distributions, it produces kernels with too small of a bandwidth (hence the over-fitting), which is likely caused by the selected number of nearest neighbors used for the localized kernel estimation being too small. Meanwhile, compared to VC, ROME is numerically more stable and does not hallucinate new modes in the estimated densities (Figure \ref{fig:baselinePlots}).
Furthermore, as part of several ablation studies, we found that ROME overcomes the shortcomings of other common density estimators, such as the over-fitting exhibited by kNN or the over-smoothing by GMM. In those studies, we additionally demonstrated that our approach of using clustering, decorrelation, and normalization is indispensable for overcoming the deficiencies of KDE.

Future work can further improve on our results by investigating the integration of more sophisticated density estimation methods, such as MPW, instead of the kernel density estimator in our proposed approach to enable better performance on non-normal clusters. 

Overall, by providing a simple way to accurately estimate distributions based on samples, ROME can help in better handling and evaluating probabilistic data as well as enabling more precise probabilistic inference.

\section*{Acknowledgments}
This research was supported by NWO-NWA project “Acting under uncertainty” (ACT), NWA.1292.19.298.

\section*{Contribution Statement}
Anna M\'esz\'aros and Julian F. Schumann equally contributed to the design and implementation of the research, to the analysis of the results and to the writing of the manuscript and should be considered joint first authors. 
Javier Alonso-Mora provided valuable feedback on the writing of the manuscript.
Arkady Zgonnikov and Jens Kober provided valuable feedback at all steps of the project and should be considered joint last authors.

\bibliographystyle{named}
\bibliography{biblio}


\clearpage

\setcounter{section}{0}
\renewcommand\thesection{\Alph{section}}
\part*{Appendix}

\section{Clustering within the OPTICS Algorithm}
In our implementation of the OPTICS algorithm [Ankerst \textit{et al.}, 1999], we follow the standard approach in regard to the generation of the reachability distances $\bm{R}_N$ and the corresponding ordered samples $\bm{X}_{I,N}$. However, we adjusted the extraction of the final set of clusters $\bm{C}$. Namely, in the standard OPTICS algorithm, one would have predetermined either the use of \textbf{DBSCAN} or $\bm{\xi}$\textbf{-clustering} together with the corresponding parameters (respectively $\epsilon$ and $\xi$). 

DBSCAN assumes that a cluster is defined by the fact that all of its samples are closer together than a predefined distance, which is the same for all clusters in a dataset. However, this assumes that clusters generally have a similar density, which might not always be the case. Therefore, $\xi$-clustering is designed to select clusters, whose member samples are significantly closer together compared to the surrounding samples, which means that this algorithm can recognize clusters with varying densities. However, it is more susceptible to noise.

Consequently, it is likely impossible to select one certain method and parameter that gives optimal performances for all potential cluster configurations. However, as this extraction is far cheaper than the previous ordering of samples, we run a large number of different cluster extractions, and use the Silhouette score to determine the optimal set of clusters. This approach makes the clustering of the samples more robust.

\section{Likelihood factors}

Given two average log likelihoods $\widehat{L}_A$ and $\widehat{L}_B$ of two density estimation methods $A$ and $B$, we can calculate a so-called \textit{factor} $\mathcal{F}$ that expresses how much more likely method $A$ is at reproducing the underlying distribution $p$ compared to method $B$:
\begin{equation*}
    \mathcal{F} = \frac{1}{100}\sum_{i=1}^{100} \mathcal{F}_i = \frac{1}{100}\sum_{i=1}^{100} \exp \left(\widehat{L}_{A,i} - \widehat{L}_{B,i}\right) \, ,
\end{equation*}
for $100$ repeated evaluations $i$. We then measure the statistic significance of $\mathcal{F} > 1$ using a standard t-test based on the $100$ $\mathcal{F}_i$ values.

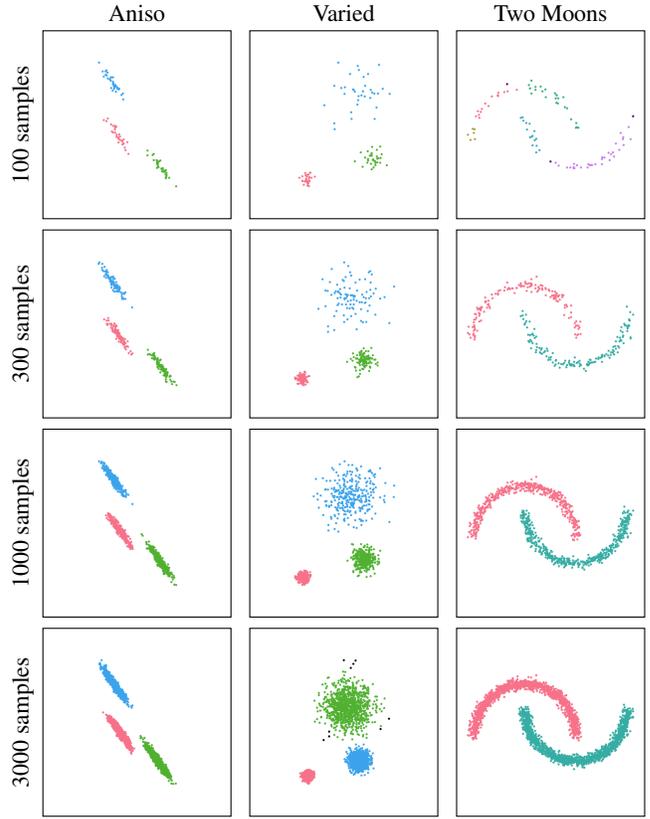
\begin{figure}[!t]
    \centering
    \begin{tikzpicture}
        \begin{scope}[xshift = 0cm]
            \node[above, rotate=90, font = \small] at (0.0, 1.25) {100 samples};
            \node[inner sep = 0] at (1.25,1.25) {\includegraphics[width = 2.5cm]{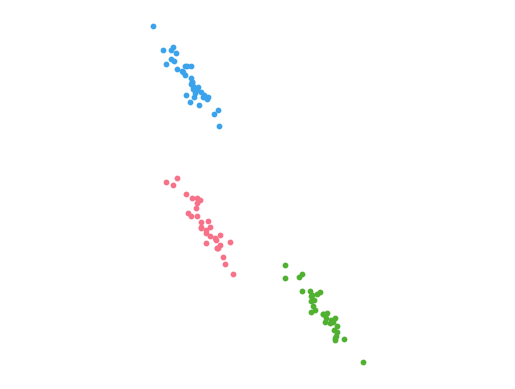}};
            \draw[black] (0,0) rectangle (2.5,2.5);
            \node[above, black, font = \small] at (1.25,2.5) {Aniso};
        \end{scope}
        \begin{scope}[xshift = 2.75cm]
            \node[inner sep = 0] at (1.25,1.25) {\includegraphics[width = 2.5cm]{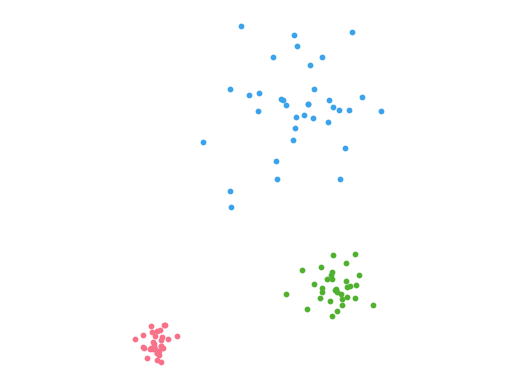}};
            \draw[black] (0,0) rectangle (2.5,2.5);
            \node[above, black, font = \small] at (1.25,2.5) {Varied};
        \end{scope}
        \begin{scope}[xshift = 5.5cm]
            \node[inner sep = 0] at (1.25,1.25) {\includegraphics[width = 2.5cm]{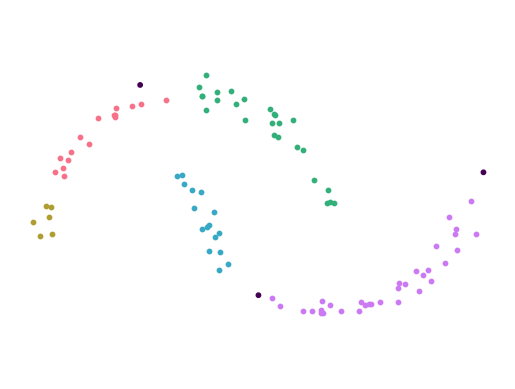}};
            \draw[black] (0,0) rectangle (2.5,2.5);
            \node[above, black, font = \small] at (1.25,2.5) {Two Moons};
        \end{scope}
        
        \begin{scope}[xshift = 0cm, yshift = -2.65cm]
            \node[above, rotate=90, font = \small] at (0.0, 1.25) {300 samples};
            \node[inner sep = 0] at (1.25,1.25) {\includegraphics[width = 2.5cm]{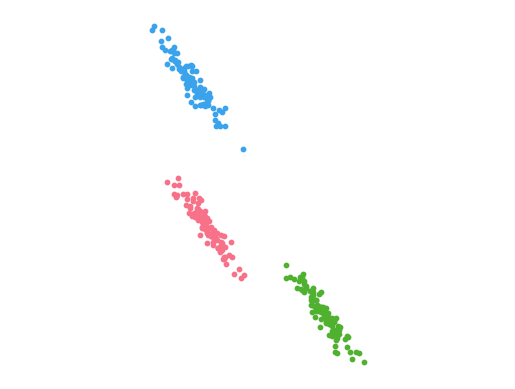}};
            \draw[black] (0,0) rectangle (2.5,2.5);
        \end{scope}
        \begin{scope}[xshift = 2.75cm, yshift = -2.65cm]
            \node[inner sep = 0] at (1.25,1.25) {\includegraphics[width = 2.5cm]{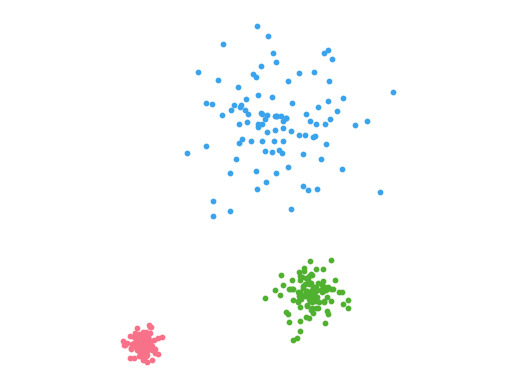}};
            \draw[black] (0,0) rectangle (2.5,2.5);
        \end{scope}
        \begin{scope}[xshift = 5.5cm, yshift = -2.65cm]
            \node[inner sep = 0] at (1.25,1.25) {\includegraphics[width = 2.5cm]{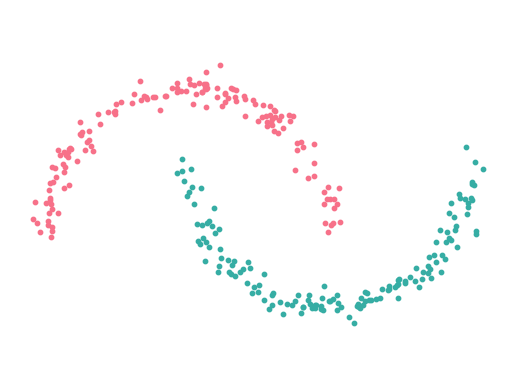}};
            \draw[black] (0,0) rectangle (2.5,2.5);
        \end{scope}
        
        \begin{scope}[xshift = 0cm, yshift = -5.3cm]
            \node[above, rotate=90, font = \small] at (0.0, 1.25) {1000 samples};
            \node[inner sep = 0] at (1.25,1.25) {\includegraphics[width = 2.5cm]{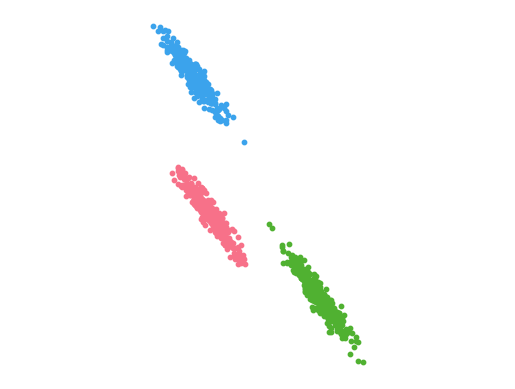}};
            \draw[black] (0,0) rectangle (2.5,2.5);
        \end{scope}
        \begin{scope}[xshift = 2.75cm, yshift = -5.3cm]
            \node[inner sep = 0] at (1.25,1.25) {\includegraphics[width = 2.5cm]{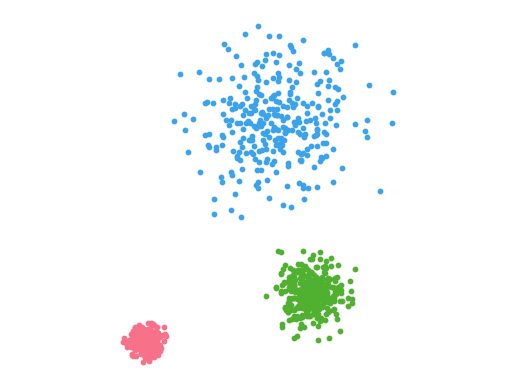}};
            \draw[black] (0,0) rectangle (2.5,2.5);
        \end{scope}
        \begin{scope}[xshift = 5.5cm, yshift = -5.3cm]
            \node[inner sep = 0] at (1.25,1.25) {\includegraphics[width = 2.5cm]{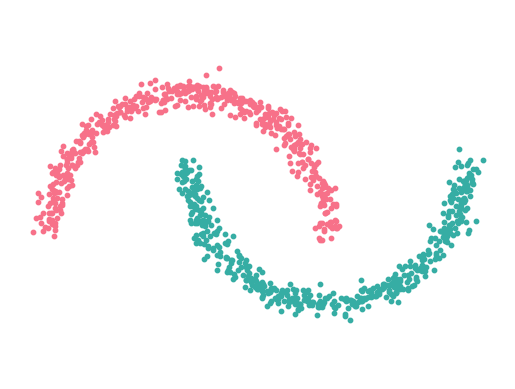}};
            \draw[black] (0,0) rectangle (2.5,2.5);
        \end{scope}
        
        \begin{scope}[xshift = 0cm, yshift = -7.95cm]
            \node[above, rotate=90, font = \small] at (0.0, 1.25) {3000 samples};
            \node[inner sep = 0] at (1.25,1.25) {\includegraphics[width = 2.5cm]{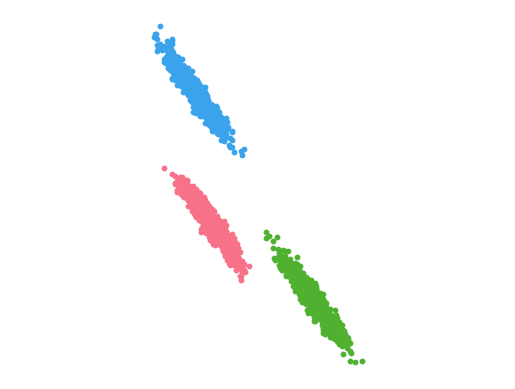}};
            \draw[black] (0,0) rectangle (2.5,2.5);
        \end{scope}
        \begin{scope}[xshift = 2.75cm, yshift = -7.95cm]
            \node[inner sep = 0] at (1.25,1.25) {\includegraphics[width = 2.5cm]{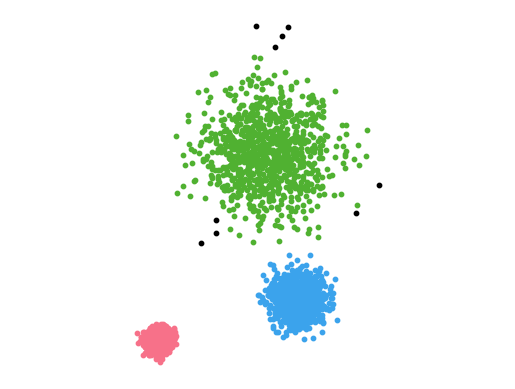}};
            \draw[black] (0,0) rectangle (2.5,2.5);
        \end{scope}
        \begin{scope}[xshift = 5.5cm, yshift = -7.95cm]
            \node[inner sep = 0] at (1.25,1.25) {\includegraphics[width = 2.5cm]{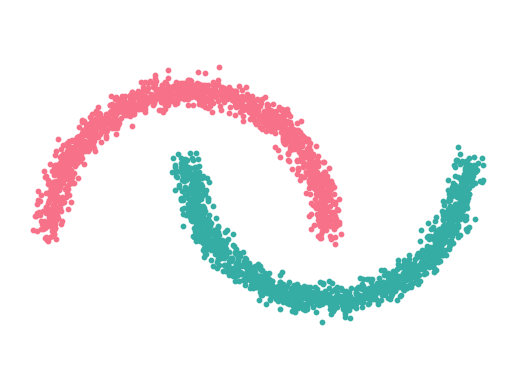}};
            \draw[black] (0,0) rectangle (2.5,2.5);
        \end{scope}
    \end{tikzpicture}
    \caption{Clustering results of OPTICS on 2D distribution datasets when using the silhouette score for the selection criterion.
    }
    \label{fig:clustering2D_silhouette}
\end{figure}

\begin{figure}[!h]
    \centering
    \begin{tikzpicture}
        \begin{scope}[xshift = 0cm]
            \node[inner sep = 0] at (1.5,1.5) {\includegraphics[width = 3cm, trim={1cm 0.5cm 1cm 0}, clip]{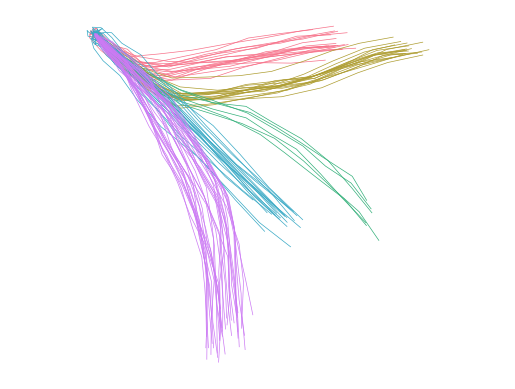}};
            \draw[black] (0,0) rectangle (3, 3);
            \node[above, black, font = \small] at (1.5,3) {Trajectories - 100 samples};
        \end{scope}
        \begin{scope}[xshift = 4cm]
            \node[inner sep = 0] at (1.5,1.5) {\includegraphics[width = 3cm, trim={1cm 0.5cm 1cm 0}, clip]{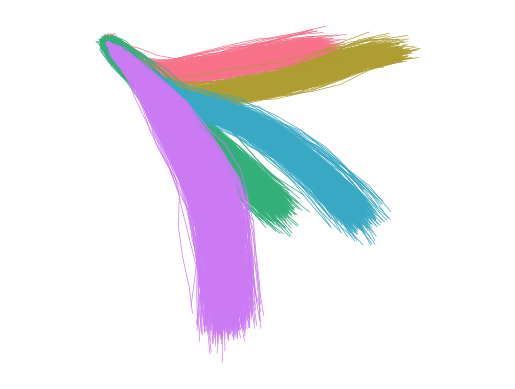}};
            \draw[black] (0,0) rectangle (3,3);
            \node[above, black, font = \small] at (1.5,3) {Trajectories - 300 samples};
        \end{scope}

        \begin{scope}[xshift = 0cm, yshift = -3.5cm]
            \node[inner sep = 0] at (1.5,1.5) {\includegraphics[width = 3cm, trim={1cm 0.5cm 1cm 0}, clip]{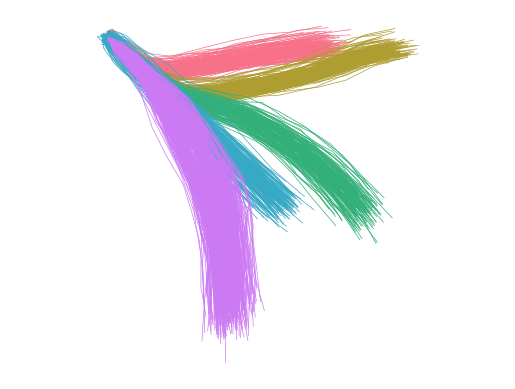}};
            \draw[black] (0,0) rectangle (3,3);
            \node[above, black, font = \small] at (1.5,3) {Trajectories - 1000 samples};
        \end{scope}
        \begin{scope}[xshift = 4cm, yshift = -3.5cm]
            \node[inner sep = 0] at (1.5,1.5) {\includegraphics[width = 3cm, trim={1cm 0.5cm 1cm 0}, clip]{Images/Clustering/Trajectories_n_samples_6000_silhouette_cluster_plot.png}};
            \draw[black] (0,0) rectangle (3,3);
            \node[above, black, font = \small] at (1.5,3) {Trajectories - 3000 samples};
        \end{scope}
        
    \end{tikzpicture}
    \caption{Clustering results of OPTICS on the Trajectories dataset when using the silhouette score for the selection criterion
    }
    \label{fig:clusteringTraj_silhouette}
\end{figure}

\section{Clustering Performance}

An important component of ROME is the OPTICS algorithm, which we utilize for identifying the clusters (i.e., the modes) within the data. 
An important aspect of OPTICS is a sensible choice of $k_{\bm{c}}$, a variable used to guard against randomness in the estimation of reachability distances (see equation (2) in the main text). In this work, we used the following rule of thumb to determine these parameters, where $N$ is the number of samples and $M$ the number of dimensions:
\begin{equation}
    k_{\bm{c}} = \min\left\{k_{\max}, \max\left\{ k_{\min}, \frac{N M}{\alpha_k}\right\} \right\} \,.
    \label{eq:kC2}
\end{equation}
This part contains three hyperparameters -- namely $k_\text{min}$, $k_\text{max}$, and $\alpha_k$.
Firstly, $k_{\min}$ is needed to ensure that the method is stable, as a too low $k_{\bm{c}}$ would make the subsequent clustering vulnerable to random sampling fluctuations. Meanwhile, $k_{\max}$ ensures that the reachability distances are actually based on only local information, and are not including points from other modes. Lastly, the term $NM / \alpha_k$ is used to ensure an independence from the number of samples, while allowing for the higher number of samples needed in higher-dimensional spaces. 

The hyperparameters were then empirically selected such that good clustering could be guaranteed over multiple datasets with a different number of samples, resulting in $k_{\min} = 5$, $k_{\max} = 20$, and $\alpha_k = 400$. With those values, we obtain satisfactory clustering performance across all datasets even when the number of samples is greatly reduced (Figures~\ref{fig:clustering2D_silhouette} and \ref{fig:clusteringTraj_silhouette}). 
The only exception to this is the Two Moons dataset with 100 samples. However, given the gaps in the intended clusters, this could be expected.

\begin{table}[!t]
\begin{tabularx}{\linewidth}{X | Y Y Y }
\toprule[1pt] 
& \multicolumn{1}{c}{$f_{\text{ROME}}$} & \multicolumn{1}{c}{$f_{\text{MPW}}$} & \multicolumn{1}{c}{$f_{\text{VC}}$}  \\ \midrule[1pt]
Aniso & {\scriptsize $\underline{0.007}\scriptscriptstyle{\pm 0.001}$} & {\scriptsize $0.016\scriptscriptstyle{\pm 0.001}$} & \textcolor{red}{\scriptsize $0.143\scriptscriptstyle{\pm 0.003}$}  \\ \midrule 
Varied & {\scriptsize $\underline{0.078}\scriptscriptstyle{\pm 0.003}$} & {\scriptsize $0.084\scriptscriptstyle{\pm 0.003}$} & {\scriptsize $0.087\scriptscriptstyle{\pm 0.003}$}  \\
\bottomrule[1pt]
\end{tabularx} 
\caption{Baseline Comparison ($D_{\text{JS}_\text{true}} \downarrow_{0}^{1}$; ground truth distributions) - marked in red is a case of notably poor performance; best values are underlined}
\label{tab: JSD_true_baseline}
\end{table}

\begin{table*}[!t]
\centering
\begin{tabularx}{\textwidth}{X | Z | Z | Z | Z | Z | Z | Z} 
\toprule[1pt] 
& \multicolumn{4}{c|}{Normalization} & \multicolumn{2}{c|}{No normalization} & \multirow{3}{*}{$f_{\text{GMM}}$} \\
& \multicolumn{2}{c|}{Decorrelation} & \multicolumn{4}{c|}{No decorrelation} & \\  
Clustering & $f_{\text{KDE}}$ & $f_{\text{kNN}}$ & $f_{\text{KDE}}$ & $f_{\text{kNN}}$ & $f_{\text{KDE}}$ & $f_{\text{kNN}}$ & \\\midrule[1pt]
Silhouette & {\cellcolor{gray!25} \scriptsize $0.007\scriptscriptstyle{\pm 0.001}$} & {\scriptsize $0.014\scriptscriptstyle{\pm 0.001}$} & {\textcolor{red}{\scriptsize $0.039\scriptscriptstyle{\pm 0.001}$}} & {\scriptsize $0.020\scriptscriptstyle{\pm 0.001}$} & {\textcolor{red}{\scriptsize $0.054\scriptscriptstyle{\pm 0.001}$}} & {\scriptsize $0.020\scriptscriptstyle{\pm 0.001}$} & {\scriptsize $0.001\scriptscriptstyle{\pm 0.000}$}  \\ \midrule 
DBCV & {\scriptsize $0.009\scriptscriptstyle{\pm 0.002}$} & {\scriptsize $0.014\scriptscriptstyle{\pm 0.001}$} & {\textcolor{red}{\scriptsize $0.037\scriptscriptstyle{\pm 0.001}$}} & {\scriptsize $0.022\scriptscriptstyle{\pm 0.001}$} & {\textcolor{red}{\scriptsize $0.061\scriptscriptstyle{\pm 0.003}$}} & {\scriptsize $0.023\scriptscriptstyle{\pm 0.002}$} & {\scriptsize $0.008\scriptscriptstyle{\pm 0.003}$}  \\ \midrule 
No clusters & {\scriptsize $0.095\scriptscriptstyle{\pm 0.001}$} & {\scriptsize $0.015\scriptscriptstyle{\pm 0.001}$} & {\scriptsize $0.245\scriptscriptstyle{\pm 0.002}$} & {\scriptsize $0.022\scriptscriptstyle{\pm 0.001}$} & {\scriptsize $0.039\scriptscriptstyle{\pm 0.001}$} & {\scriptsize $0.023\scriptscriptstyle{\pm 0.001}$} & {\scriptsize $0.530\scriptscriptstyle{\pm 0.002}$}  \\ \bottomrule[1pt]
\end{tabularx} 
\caption{Ablations ($D_{\text{JS}_\text{true}} \downarrow_{0}^{1}$, Aniso; comparison to the ground truth distribution) - When clustering, decorrelation and
normalization improve results for distributions with high intra-mode
correlation. ROME highlighted in gray.}
\label{tab: JSD_true_aniso}
\end{table*}

\begin{table*}[!t]
\centering
\begin{tabularx}{\textwidth}{X | Z | Z | Z | Z | Z | Z | Z} 
\toprule[1pt] 
& \multicolumn{4}{c|}{Normalization} & \multicolumn{2}{c|}{No normalization} & \multirow{3}{*}{$f_{\text{GMM}}$} \\
& \multicolumn{2}{c|}{Decorrelation} & \multicolumn{4}{c|}{No decorrelation} & \\  
Clustering & $f_{\text{KDE}}$ & $f_{\text{kNN}}$ & $f_{\text{KDE}}$ & $f_{\text{kNN}}$ & $f_{\text{KDE}}$ & $f_{\text{kNN}}$ & \\\midrule[1pt]
Silhouette & {\cellcolor{gray!25} \scriptsize $0.078\scriptscriptstyle{\pm 0.003}$} & {\scriptsize $0.083\scriptscriptstyle{\pm 0.003}$} & {\scriptsize $0.078\scriptscriptstyle{\pm 0.003}$} & {\scriptsize $0.083\scriptscriptstyle{\pm 0.003}$} & {\scriptsize $0.070\scriptscriptstyle{\pm 0.003}$} & {\scriptsize $0.083\scriptscriptstyle{\pm 0.003}$} & {\scriptsize $0.074\scriptscriptstyle{\pm 0.003}$}  \\ \midrule 
DBCV & {\scriptsize $0.074\scriptscriptstyle{\pm 0.003}$} & {\scriptsize $0.084\scriptscriptstyle{\pm 0.003}$} & {\scriptsize $0.074\scriptscriptstyle{\pm 0.003}$} & {\scriptsize $0.084\scriptscriptstyle{\pm 0.003}$} & {\scriptsize $0.065\scriptscriptstyle{\pm 0.003}$} & {\scriptsize $0.085\scriptscriptstyle{\pm 0.003}$} & {\scriptsize $0.068\scriptscriptstyle{\pm 0.003}$}  \\ \midrule 
No clusters & {\textcolor{red}{\scriptsize $0.173\scriptscriptstyle{\pm 0.004}$}} & {\scriptsize $0.081\scriptscriptstyle{\pm 0.003}$} & {\textcolor{red}{\scriptsize $0.183\scriptscriptstyle{\pm 0.004}$}} & {\scriptsize $0.080\scriptscriptstyle{\pm 0.003}$} & {\scriptsize $0.074\scriptscriptstyle{\pm 0.003}$} & {\scriptsize $0.080\scriptscriptstyle{\pm 0.003}$} & {\textcolor{red}{\scriptsize $0.540\scriptscriptstyle{\pm 0.003}$}}  \\ \bottomrule[1pt]
\end{tabularx} 
\caption{Ablations ($D_{\text{JS}_\text{true}} \downarrow_{0}^{1}$, Varied; comparison to the ground truth distribution) - Clustering is essential to improving results for distributions with varying mode densities. ROME highlighted in gray.}
\label{tab: JSD_true_varied}
\end{table*}

\begin{table*}[!t]
    \centering
    \begin{tabularx}{\textwidth}{X | Y Y Y | Y Y Y | Y Y Y}
    \toprule[1pt] 
    & \multicolumn{3}{c|}{$D_\text{JS} \downarrow_{0}^{1}$} & \multicolumn{3}{c|}{$\widehat{W} \rightarrow 0$} & \multicolumn{3}{c}{$\widehat{L} \uparrow$}  \\& \multicolumn{1}{c}{$f_{\text{ROME}}$} & \multicolumn{1}{c}{$f_{\text{MPW}}$} & \multicolumn{1}{c|}{$f_{\text{VC}}$} & \multicolumn{1}{c}{$f_{\text{ROME}}$} & \multicolumn{1}{c}{$f_{\text{MPW}}$} & \multicolumn{1}{c|}{$f_{\text{VC}}$} & \multicolumn{1}{c}{$f_{\text{ROME}}$} & \multicolumn{1}{c}{$f_{\text{MPW}}$} & \multicolumn{1}{c}{$f_{\text{VC}}$}  \\ \midrule[1pt]
    Gaussian & {\scriptsize $\underline{0.006}\scriptscriptstyle{\pm 0.001}$} & {\scriptsize $0.023\scriptscriptstyle{\pm 0.002}$} & {\scriptsize $0.007\scriptscriptstyle{\pm 0.001}$} & {\scriptsize $-0.02\scriptscriptstyle{\pm 0.09}$} & {\scriptsize $-0.26\scriptscriptstyle{\pm 0.05}$} & {\scriptsize $\underline{-0.01}\scriptscriptstyle{\pm 0.11}$} & {\scriptsize $\underline{-2.84}\scriptscriptstyle{\pm 0.02}$} & {\scriptsize $-2.87\scriptscriptstyle{\pm 0.02}$} & {\scriptsize $-2.85\scriptscriptstyle{\pm 0.02}$}  \\ \midrule 
    Elliptical & {\scriptsize $\underline{0.005}\scriptscriptstyle{\pm 0.001}$} & {\scriptsize $0.023\scriptscriptstyle{\pm 0.002}$} & {\scriptsize $0.007\scriptscriptstyle{\pm 0.001}$} & {\scriptsize $0.05\scriptscriptstyle{\pm 0.17}$} & {\scriptsize $-0.29\scriptscriptstyle{\pm 0.08}$} & {\scriptsize $\underline{0.04}\scriptscriptstyle{\pm 0.16}$} & {\scriptsize $\underline{-1.64}\scriptscriptstyle{\pm 0.02}$} & {\scriptsize $-1.66\scriptscriptstyle{\pm 0.02}$} & {\scriptsize $\underline{-1.64}\scriptscriptstyle{\pm 0.02}$}  \\ \midrule 
    Rot. Ellip. & {\scriptsize $0.005\scriptscriptstyle{\pm 0.001}$} & {\scriptsize $0.023\scriptscriptstyle{\pm 0.002}$} & {\scriptsize $\underline{0.003}\scriptscriptstyle{\pm 0.001}$} & {\scriptsize $\underline{0.04}\scriptscriptstyle{\pm 0.20}$} & {\scriptsize $-0.35\scriptscriptstyle{\pm 0.11}$} & {\scriptsize $0.84\scriptscriptstyle{\pm 0.32}$} & {\scriptsize $\underline{-1.42}\scriptscriptstyle{\pm 0.02}$} & {\scriptsize $-1.45\scriptscriptstyle{\pm 0.02}$} & {\scriptsize $-1.55\scriptscriptstyle{\pm 0.01}$}  \\ \midrule 
    Uni. Traj. & {\scriptsize $\underline{0.001}\scriptscriptstyle{\pm 0.000}$} & {\scriptsize $0.003\scriptscriptstyle{\pm 0.000}$} & {\textcolor{red}{\scriptsize $0.801\scriptscriptstyle{\pm 0.003}$}} & {\scriptsize $1.69\scriptscriptstyle{\pm 0.01}$} & {\scriptsize $2.33\scriptscriptstyle{\pm 0.01}$} & {\scriptsize $\underline{1.60}\scriptscriptstyle{\pm 0.11}$} & {\scriptsize $\underline{32.46}\scriptscriptstyle{\pm 0.02}$} & {\scriptsize $28.20\scriptscriptstyle{\pm 0.02}$} & {\textcolor{red}{\scriptsize $-193.88\scriptscriptstyle{\pm 17.5}$}}  \\ \bottomrule[1pt]
    \end{tabularx} 
    \caption{Baseline Comparison, uni-modal -- marked in red are cases with notably poor performance; best values are underlined.}    
    \label{tab:baselines_uniModal}

\end{table*}

\begin{table}
    \centering
    \begin{tabularx}{\linewidth}{X | Y Y Y}
    \toprule[1pt] 
    & \multicolumn{3}{c}{$D_{\text{JS}_\text{true}} \downarrow_{0}^{1}$}  \\
    & \multicolumn{1}{c}{$f_{\text{ROME}}$} & \multicolumn{1}{c}{$f_{\text{MPW}}$} & \multicolumn{1}{c}{$f_{\text{VC}}$}  \\ \midrule[1pt]
    Gaussian & {\scriptsize $\underline{0.004}\scriptscriptstyle{\pm 0.001}$} & {\scriptsize $0.012\scriptscriptstyle{\pm 0.001}$} & {\scriptsize $\underline{0.004}\scriptscriptstyle{\pm 0.001}$} \\  \midrule 
    Elliptical & {\scriptsize $\underline{0.004}\scriptscriptstyle{\pm 0.001}$} & {\scriptsize $0.012\scriptscriptstyle{\pm 0.001}$} & {\scriptsize $\underline{0.004}\scriptscriptstyle{\pm 0.001}$}  \\  \midrule
    Rot. Elliptical & {\scriptsize $\underline{0.004}\scriptscriptstyle{\pm 0.001}$} & {\scriptsize $0.013\scriptscriptstyle{\pm 0.001}$} & {\scriptsize $0.029\scriptscriptstyle{\pm 0.001}$} \\ \bottomrule[1pt]
    \end{tabularx} 
    \caption{Baseline Comparison ($D_{\text{JS}_\text{true}} \downarrow_{0}^{1}$; ground truth uni-modal distributions) - best values are underlined.}
    \label{tab:baseline_uniMod_true}
\end{table}
\section{Comparison to Ground Truth Distributions}

In order to get a better idea of the goodness of fit of the different density estimators, we also compare the distributions $\widehat{p}_{1} = f(\bm{X}_1)$ obtained from the estimator $f$ to the ground truth distribution $p$ using the Jensen-Shannon divergence $D_\text{JS}(\widehat{p}_{1}\Vert p)$. To simplify notation, this metric will be referred to as $D_{\text{JS}_\text{true}}$. The ground truth distribution $p$ is only available for Aniso and Varied.

\begin{figure}[!h]
    \centering
    \begin{tikzpicture}
        \begin{scope}[xshift = 0cm]
            \node[above right, inner sep = 0] at (0,0) {\includegraphics[width = 2.5cm]{Images/aniso_ROME_samples.png}};
            \draw[black] (0,0) rectangle (2.5,2.5);
            \node[above, black, font = \small] at (1.25,2.5) {ROME};
        \end{scope}
        \begin{scope}[xshift = 3cm]
            \node[above right, inner sep = 0] at (0,0) {\includegraphics[width = 2.5cm]{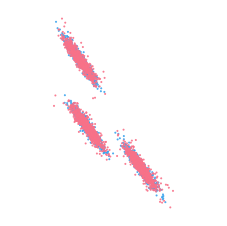}};
            \draw[black] (0,0) rectangle (2.5,2.5);
            \node[above, black, font = \small] at (1.25,2.5) {MPW};
        \end{scope}
        \begin{scope}[xshift = 6cm]
            \node[above right, inner sep = 0] at (0,0) {\includegraphics[width = 2.5cm]{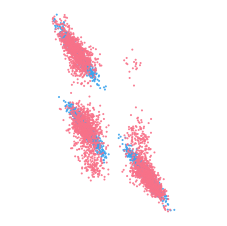}};
            \draw[black] (0,0) rectangle (2.5,2.5);
            \node[above, black, font = \small] at (1.25,2.5) {VC};
        \end{scope}
    \end{tikzpicture}
    \caption{Samples obtained with ROME, MPW and VC (pink) contrasted with samples from $p$ (blue); Aniso. 
    }
    \label{fig:baselinePlots_aniso}
\end{figure}

In the baseline comparison we observe that our estimator ROME achieves the lowest $D_{\text{JS}_\text{true}}$ values, particularly in the case of the Aniso distribution (Table~\ref{tab: JSD_true_baseline}). We further see that $f_\text{VC}$ is unable to fit the Aniso dataset well, which is supported by further visual inspection (Figure~\ref{fig:baselinePlots_aniso}). 

Within our ablation study on the Aniso dataset we clearly observe the importance of both decorrelation and normalisation in the case of highly correlated features (Table~\ref{tab: JSD_true_aniso}). 
Meanwhile, in our ablation study on the Varied dataset, we continue to observe the importance of clustering for a good fit on data with varying densities across modes, as indicated by the high $D_{\text{JS}_\text{true}}$ values when no clustering is used (Table~\ref{tab: JSD_true_varied}).
In the ablation cases, as we cannot guarantee that using $f_{\text{kNN}}$ results in a normalized distribution, the corresponding values should be considered with caution.

\section{Comparison on Uni-modal Distribution}
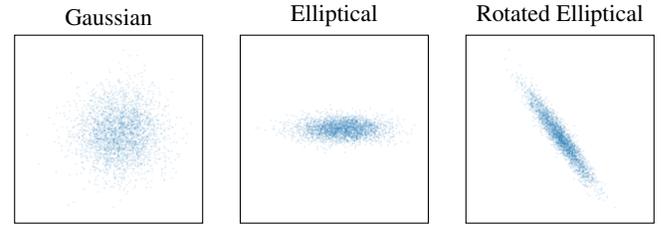
\begin{figure}[!h]
    \centering
    \begin{tikzpicture}
        \begin{scope}[xshift = 0cm]
            \node[above right, inner sep = 0] at (0,0) {\includegraphics[width = 2.5cm]{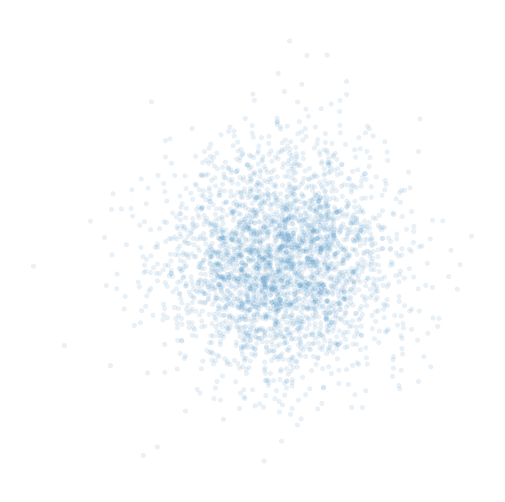}};
            \draw[black] (0,0) rectangle (2.5,2.5);
            \node[above, black, font = \small] at (1.25,2.5) {Gaussian};
        \end{scope}
        \begin{scope}[xshift = 3cm]
            \node[above right, inner sep = 0] at (0,0) {\includegraphics[width = 2.5cm]{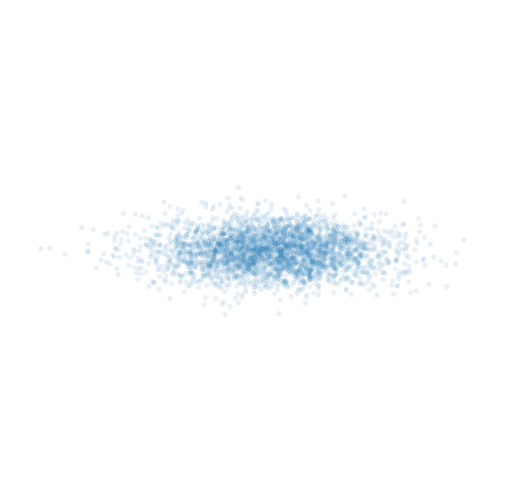}};
            \draw[black] (0,0) rectangle (2.5,2.5);
            \node[above, black, font = \small] at (1.25,2.5) {Elliptical};
        \end{scope}
        \begin{scope}[xshift = 6cm]
            \node[above right, inner sep = 0] at (0,0) {\includegraphics[width = 2.5cm]{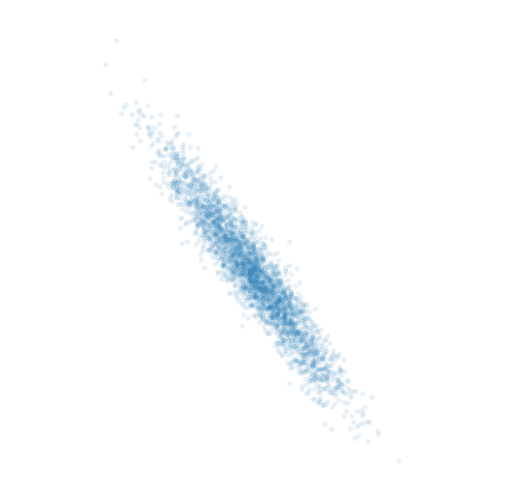}};
            \draw[black] (0,0) rectangle (2.5,2.5);
            \node[above, black, font = \small, align=center] at (1.25,2.5) {Rotated Elliptical};
        \end{scope}
    \end{tikzpicture}
    \caption{Samples from the two-dimensional uni-modal distributions. 
    }
    \label{fig:unimodal_2D}
\end{figure}

\begin{figure}[!h]
    \centering
    \begin{tikzpicture}
        \node[above right, inner sep = 0] at (0,-2) {\includegraphics[width = 5cm,clip]{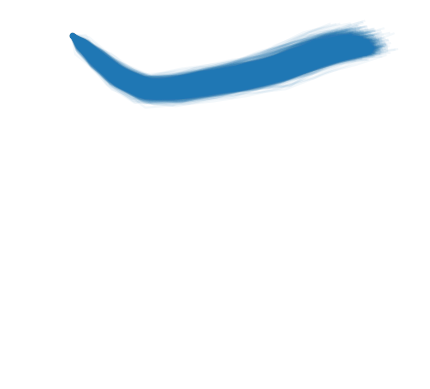}};
        \begin{axis}[ 
            at = {(0cm, 0cm)}, 
            width = 5cm, 
            height = 2.500cm, 
            scale only axis = true, 
            axis lines = left, 
            xmin = 3, 
            xmax = 8, 
            xtick = {3, 4, 5, 6, 7}, 
            xticklabels = {3, 4, 5, 6, 7}, 
            x tick label style = {font=\scriptsize}, 
            xmajorgrids = false, 
            ymin = 6, 
            ymax = 8.5, 
            ytick = {6, 7, 8}, 
            yticklabels = {6, 7, 8}, 
            y tick label style = {font=\scriptsize, rotate = 90}, 
            ymajorgrids = false, 
        ] 
        \end{axis}
        \node[below, black, font = \small] at (2.5,-0.4) {$x$ [$m$]};
        \node[above, black, font = \small, rotate = 90] at (-0.4, 1.25) {$y$ [$m$]};
    \end{tikzpicture}
    \vspace{-1.3cm}
    \caption{Samples from the uni-modal pedestrian trajectory distribution; obtained by selecting a single mode from the original multi-modal distribution.}
    \label{fig:unimodal_trajectories}
\end{figure}

In order to verify that ROME is not tailored only to multi-modal distributions, but can in fact be applied to simpler uni-modal distributions as well, we conduct experiments on the Gaussian, Elliptical and Rotated Elliptical distributions (Figure~\ref{fig:unimodal_2D}), as well as on a more complex multivariate distribution using a single mode of our Trajectories dataset (Figure~\ref{fig:unimodal_trajectories}). 
We perform both the baseline comparison as well as the ablation study as done for the multi-modal datasets presented in the main body of the paper.

Within the baseline comparison, ROME continues to achieve high performance compared to the baselines across all three metrics on the uni-modal datasets (Table~\ref{tab:baselines_uniModal}) and are further supported by the results of $D_{\text{JS}_\text{true}}$ on the ground truth distributions of the two-dimensional uni-modal distributions (Table~\ref{tab:baseline_uniMod_true}). 
The only exception is the Wasserstein-based metric $\widehat{W}$, where VC achieves slightly better performance.
However, when this information is coupled with the values of $D_{\text{JS}}$ and $\widehat{L}$, we can conclude that VC struggles in fitting distributions with strongly correlated features as seen by the results on the Rotated Elliptical distribution and the multivariate Trajectories distribution.

Within the ablation study, it is evident that clustering has little to no effect on the final density estimation, which is to be expected since there is only a single cluster in these datasets. 
Unsurprisingly, in the case of the two-dimensional uni-modal distributions, using $f_\text{GMM}$ achieves the best results since the underlying distributions are Gaussian by nature. However, ROME still leads to a good fit on these simple distributions.
The ablations on the Elliptical and Rotated Elliptical distributions once more demonstrate the importance of normalizing as well as decorrelating the data prior to the density estimation, particularly when using $f_\text{KDE}$ as the underlying density estimator (Tables~\ref{tab:aniso_uniMod_JSDtrue} and \ref{tab:anisoRot_uniMod_JSDtrue}).

Finally, the results on the uni-modal Trajectories dataset exhibit similar trends to the ones on the four multi-modal datasets presented in the main body of the paper. 
These trends include a tendency towards over-smoothing in the case of using $f_\text{GMM}$ as the underlying density estimator in the case of non-normal distributions (Table~\ref{tab:trajectories_uniMod_wasserstein}).
Additionally, we observe that normalizing and decorrelating the data prior to performing the density estimation plays an important role when using $f_\text{KDE}$ as the underlying density estimator -- which was selected for ROME (Tables~\ref{tab:trajectories_uniMod_JSD}-\ref{tab:trajectories_uniMod_wasserstein}).

\newpage

\begin{table}[!t]
    \centering
    \begin{tabularx}{\linewidth}{X | Z | Z | Z | Z } 
    \toprule[1pt] 
    & \multicolumn{2}{c|}{Norm.} & \multicolumn{1}{c|}{No norm.} & \multirow{2}{*}{$f_{\text{GMM}}$} \\
    Clustering & \multicolumn{1}{c|}{Decorr.} & \multicolumn{2}{c|}{No decorr.} & \\  \midrule[1pt]
    Silhouette & {\cellcolor{gray!25}\scriptsize $0.004\scriptscriptstyle{\pm 0.001}$} & {\scriptsize $0.004\scriptscriptstyle{\pm 0.001}$} & \textcolor{red}{\scriptsize $0.017\scriptscriptstyle{\pm 0.001}$} & {\scriptsize $0.000\scriptscriptstyle{\pm 0.000}$}  \\ \midrule 
    DBCV & {\scriptsize $0.004\scriptscriptstyle{\pm 0.001}$} & {\scriptsize $0.004\scriptscriptstyle{\pm 0.001}$} & \textcolor{red}{\scriptsize $0.017\scriptscriptstyle{\pm 0.001}$} & {\scriptsize $0.000\scriptscriptstyle{\pm 0.000}$}  \\ \midrule 
    No clusters & {\scriptsize $0.004\scriptscriptstyle{\pm 0.001}$}  & {\scriptsize $0.004\scriptscriptstyle{\pm 0.001}$} & \textcolor{red}{\scriptsize $0.017\scriptscriptstyle{\pm 0.001}$} & {\scriptsize $0.000\scriptscriptstyle{\pm 0.000}$}  \\ \bottomrule[1pt]
    \end{tabularx} 
    \caption{Ablations ($D_{\text{JS}_\text{true}} \downarrow_{0}^{1}$, Elliptical) - Normalization improves the fit for anisotropic Gaussian distributions. ROME highlighted in gray.}    
    \label{tab:aniso_uniMod_JSDtrue}
\end{table}

\begin{table}[!t]
    \centering
    \begin{tabularx}{\linewidth}{X | Z | Z | Z | Z } 
    \toprule[1pt] 
    & \multicolumn{2}{c|}{Norm.} & \multicolumn{1}{c|}{No norm.} & \multirow{2}{*}{$f_{\text{GMM}}$} \\
    Clustering & \multicolumn{1}{c|}{Decorr.} & \multicolumn{2}{c|}{No decorr.} & \\  \midrule[1pt]
    Silhouette & {\cellcolor{gray!25}\scriptsize $0.004\scriptscriptstyle{\pm 0.001}$} & \textcolor{red}{\scriptsize $0.027\scriptscriptstyle{\pm 0.001}$} & \textcolor{red}{\scriptsize $0.038\scriptscriptstyle{\pm 0.001}$} & {\scriptsize $0.000\scriptscriptstyle{\pm 0.000}$}  \\ \midrule 
    DBCV & {\scriptsize $0.004\scriptscriptstyle{\pm 0.001}$} & \textcolor{red}{\scriptsize $0.027\scriptscriptstyle{\pm 0.001}$} & \textcolor{red}{\scriptsize $0.038\scriptscriptstyle{\pm 0.001}$} & {\scriptsize $0.000\scriptscriptstyle{\pm 0.000}$}  \\ \midrule 
    No clusters & {\scriptsize $0.004\scriptscriptstyle{\pm 0.001}$} & \textcolor{red}{\scriptsize $0.027\scriptscriptstyle{\pm 0.001}$} & \textcolor{red}{\scriptsize $0.038\scriptscriptstyle{\pm 0.001}$} & {\scriptsize $0.000\scriptscriptstyle{\pm 0.000}$}  \\ \bottomrule[1pt]
    \end{tabularx} 
    
    \caption{Ablations ($D_{\text{JS}_\text{true}} \downarrow_{0}^{1}$, Rotated Elliptical) - Normalization and decorrelation improves the fit for distributions with strongly correlated features. ROME highlighted in gray.}    
    \label{tab:anisoRot_uniMod_JSDtrue}
\end{table}

\begin{table}[!t]
    \centering
    \begin{tabularx}{\linewidth}{X | Z | Z | Z | Z } 
    \toprule[1pt] 
    & \multicolumn{2}{c|}{Norm.} & \multicolumn{1}{c|}{No norm.} & \multirow{2}{*}{$f_{\text{GMM}}$} \\
    Clustering & \multicolumn{1}{c|}{Decorr.} & \multicolumn{2}{c|}{No decorr.} & \\  \midrule[1pt]
Silhouette & {\cellcolor{gray!25} \scriptsize $\hphantom{0}32.46\scriptscriptstyle{\pm 0.02}$} & {\scriptsize $\hphantom{0}31.87\scriptscriptstyle{\pm 0.10}$} & \textcolor{red}{\scriptsize $-13.85\scriptscriptstyle{\pm 0.00}$} &{\scriptsize $\hphantom{0}28.61\scriptscriptstyle{\pm 0.02}$}  \\ \midrule 
DBCV & {\scriptsize $\hphantom{0}32.46\scriptscriptstyle{\pm 0.02}$} & {\scriptsize $\hphantom{0}31.87\scriptscriptstyle{\pm 0.10}$} & \textcolor{red}{\scriptsize $-14.51\scriptscriptstyle{\pm 1.96}$} & {\scriptsize $\hphantom{0}28.61\scriptscriptstyle{\pm 0.02}$}  \\ \midrule 
No clusters & {\scriptsize $\hphantom{0}32.46\scriptscriptstyle{\pm 0.02}$} & {\scriptsize $\hphantom{0}31.87\scriptscriptstyle{\pm 0.10}$} & \textcolor{red}{\scriptsize $-13.85\scriptscriptstyle{\pm 0.00}$} & {\scriptsize $\hphantom{0}28.61\scriptscriptstyle{\pm 0.02}$}  \\ \bottomrule[1pt]
\end{tabularx} 
    \caption{Ablations ($\widehat{L} \uparrow$, Uni-modal Trajectories) - Excluding normalization
 is not robust against non-normal distributions. ROME highlighted in gray.}
    \label{tab:trajectories_uniMod_likelihood}
\end{table}

\begin{table}[!t]
    \centering
    \begin{tabularx}{\linewidth}{X | Z | Z | Z | Z } 
    \toprule[1pt] 
    & \multicolumn{2}{c|}{Norm.} & \multicolumn{1}{c|}{No norm.} & \multirow{2}{*}{$f_{\text{GMM}}$} \\
    Clustering & \multicolumn{1}{c|}{Decorr.} & \multicolumn{2}{c|}{No decorr.} & \\  \midrule[1pt]
    Silhouette & {\cellcolor{gray!25} \scriptsize $0.013\scriptscriptstyle{\pm 0.004}$} & \textcolor{red}{\scriptsize $0.166\scriptscriptstyle{\pm 0.063}$} & {\scriptsize $0.000\scriptscriptstyle{\pm 0.000}$} & {\scriptsize $0.005\scriptscriptstyle{\pm 0.003}$}  \\ \midrule 
    DBCV & {\scriptsize $0.013\scriptscriptstyle{\pm 0.004}$} & \textcolor{red}{\scriptsize $0.166\scriptscriptstyle{\pm 0.063}$} & {\scriptsize $0.005\scriptscriptstyle{\pm 0.031}$} & {\scriptsize $0.005\scriptscriptstyle{\pm 0.003}$}  \\ \midrule 
    No clusters & {\scriptsize $0.013\scriptscriptstyle{\pm 0.004}$} & \textcolor{red}{\scriptsize $0.166\scriptscriptstyle{\pm 0.063}$} & {\scriptsize $0.000\scriptscriptstyle{\pm 0.000}$} & {\scriptsize $0.005\scriptscriptstyle{\pm 0.003}$}  \\ \bottomrule[1pt]
    \end{tabularx} 
    \caption{Ablations ($D_\text{JS} \downarrow_{0}^{1}$, Uni-modal Trajectories) -  Using decorrelation is necessary to avoid overfitting when using normalization.}    
    \label{tab:trajectories_uniMod_JSD}
\end{table}

\begin{table}[!t]
    \centering
    \begin{tabularx}{\linewidth}{X | Z | Z | Z | Z} 
    \toprule[1pt] 
    & \multicolumn{2}{c|}{Norm.} & \multicolumn{1}{c|}{No norm.} & \multirow{2}{*}{$f_{\text{GMM}}$} \\
    Clustering & \multicolumn{1}{c|}{Decorr.} & \multicolumn{2}{c|}{No decorr.} & \\ \midrule[1pt]
    Silhouette & {\cellcolor{gray!25} \scriptsize $\hphantom{0}1.69\scriptscriptstyle{\pm 0.01}$} &  {\scriptsize $\hphantom{0}1.68\scriptscriptstyle{\pm 0.02}$} &  \textcolor{red}{\scriptsize $22.39\scriptscriptstyle{\pm 0.10}$} &  \textcolor{red}{\scriptsize $\hphantom{0}2.20\scriptscriptstyle{\pm 0.03}$}  \\ \midrule 
    DBCV & {\scriptsize $\hphantom{0}1.69\scriptscriptstyle{\pm 0.01}$} &  {\scriptsize $\hphantom{0}1.68\scriptscriptstyle{\pm 0.02}$} &  \textcolor{red}{\scriptsize $23.18\scriptscriptstyle{\pm 2.38}$} &  \textcolor{red}{\scriptsize $\hphantom{0}2.20\scriptscriptstyle{\pm 0.03}$}  \\ \midrule 
    No clusters & {\scriptsize $\hphantom{0}1.69\scriptscriptstyle{\pm 0.01}$} &  {\scriptsize $\hphantom{0}1.68\scriptscriptstyle{\pm 0.02}$} &  \textcolor{red}{\scriptsize $22.39\scriptscriptstyle{\pm 0.09}$} &  \textcolor{red}{\scriptsize $\hphantom{0}2.20\scriptscriptstyle{\pm 0.03}$}  \\ \bottomrule[1pt]
    \end{tabularx} 
   \caption{Ablations ($\widehat{W} \rightarrow 0$, Uni-modal Trajectories) - Excluding normalization
or using $f_\text{GMM}$ as the downstream estimator is not robust against non-normal distributions. ROME highlighted in gray.}
\label{tab:trajectories_uniMod_wasserstein}
\end{table}

\end{document}